\documentclass[conference]{IEEEtran}
\IEEEoverridecommandlockouts
\usepackage{cite}
\usepackage{amsmath,amssymb,amsfonts}
\usepackage{algorithmic}
\usepackage{graphicx}
\usepackage{textcomp}
\usepackage{xcolor}
\def\BibTeX{{\rm B\kern-.05em{\sc i\kern-.025em b}\kern-.08em
    T\kern-.1667em\lower.7ex\hbox{E}\kern-.125emX}}

\usepackage{booktabs} 
\usepackage{epstopdf}
\usepackage[linesnumbered,vlined,ruled]{algorithm2e}

\newtheorem{theo}{Lemma}

\newtheorem{defi}{Definition}

 \SetKwProg{Fn}{Function}{}{}
 
\begin{document}

\title{A Structural Approach to Dynamic Migration in Petri Net Models of Structured Workflows}

\author{\IEEEauthorblockN{Ahana Pradhan*\thanks{*Author is currently affiliated to Huawei Technologies India Pvt. Ltd.} and Rushikesh K. Joshi}
\IEEEauthorblockA{\textit{Department of Computer Science \& Technology} \\
\textit{Indian Institute of Technology Bombay}\\
Mumbai, India \\
rkj@cse.iitb.ac.in}
}

\maketitle

\begin{abstract}
In the context of dynamic evolution of workflow processes, the 
{\em change region} identifies the part of the old process from which migration 
to the new process is guaranteed to be inconsistent. However, this
approach may lead to
overestimated regions, incorrectly identifying migratable instances as 
non-migratable. This overestimation causes delays due to postponement of 
immediate migration. The paper analyzes 
this overestimation problem on a class of Petri nets models. 
Structural properties leading to  
conditions for minimal change regions and overestimations are developed 
resulting into classification of change regions into two types of change regions called Structural Change Regions and Perfect Structural Change Regions. Necessary and sufficient conditions for perfect regions are identified. The paper also discusses ways for computing the same in terms of structural properties of the old and the new processes.
\end{abstract}


\section{Introduction}

\textit{Business processes} represent complex flows of tasks achieving 
business goals of organizations. Examples of software assisted business 
processes are those for opening bank accounts, for obtaining passports and 
visas, and for obtaining travel bookings. At any point of time, many instances 
of such processes may be active in the business.
 Use of process aware information 
systems is aimed at offloading much 
of the manual operations of process creation, instance creation, monitoring, 
resource 
management, exception handling and design improvements.
Examples of such systems are the BPM system from Oracle 
\cite{oracle}, jBPM from JBoss \cite{jbpm}, and YAWL \cite{ter2009modern}. 

In such process automation systems, \textit{dynamic evolution} of a process 
becomes a challenge when many instances of the evolving process are active and 
partially completed. The most significant challenge faced by such 
automation is to compute a consistent state of the instance in the new process 
such that the new state (state after migration) is acceptable in the business 
process 
as equivalent to the old one. If such a migration is not possible at a 
particular state of the instance, the migration is held up till such a state is 
reached.  This problem of identifying \textit{non-migratable} states is 
addressed by identifying the regions in the old net, which cover all the 
non-migratable states. These regions are called \textit{change regions}, which 
were 
conceptualized in the works of Ellis et al. \cite{ellis1995dynamic} and Van der Aalst 
\cite{van2001exterminating}. States of the instances that do not overlap with 
the change region are migratable, and are non-migratable otherwise. 
The earlier approaches to compute change regions
\cite{van2001exterminating}, \cite{Sun2009284} do not result in false-positives 
that are underestimates, which identify an unsafe regions to be safe. However, 
as noted in \cite{van2001exterminating}, and \cite{Cicirelli20101148}, they do 
suffer from false-negatives that  are overestimates, which identify a safe 
region to be unsafe.

In this paper,  we explore this notion of  {\em change 
region} further by showing the cases in which it can be optimally computed 
without incurring false negatives. Our work 
carries out an extensive characterization of the notion of change region in the 
context of dynamic workflow evolution,  coming up with identification of two 
types of 
change regions called \textit{Structural Change Regions} (SCR) and 
\textit{Perfect Structural Change 
Regions} (PSCR). 
The characterization of 
the two type of change regions and their computations
are developed  in terms of set operations
based on a series of structural properties and lemmas. 
However, it is shown that PSCR may not exist in certain 
change situations. The necessary and sufficient 
conditions for the existence of the PSCR are identified. 
The approach of PSCR is also compared with the existing change 
region approach in the literature.

\subsection{Preliminaries: Petri Nets for Modeling Workflows}

\begin{figure*}[h!bt]
\centering

 \includegraphics[scale=0.6]{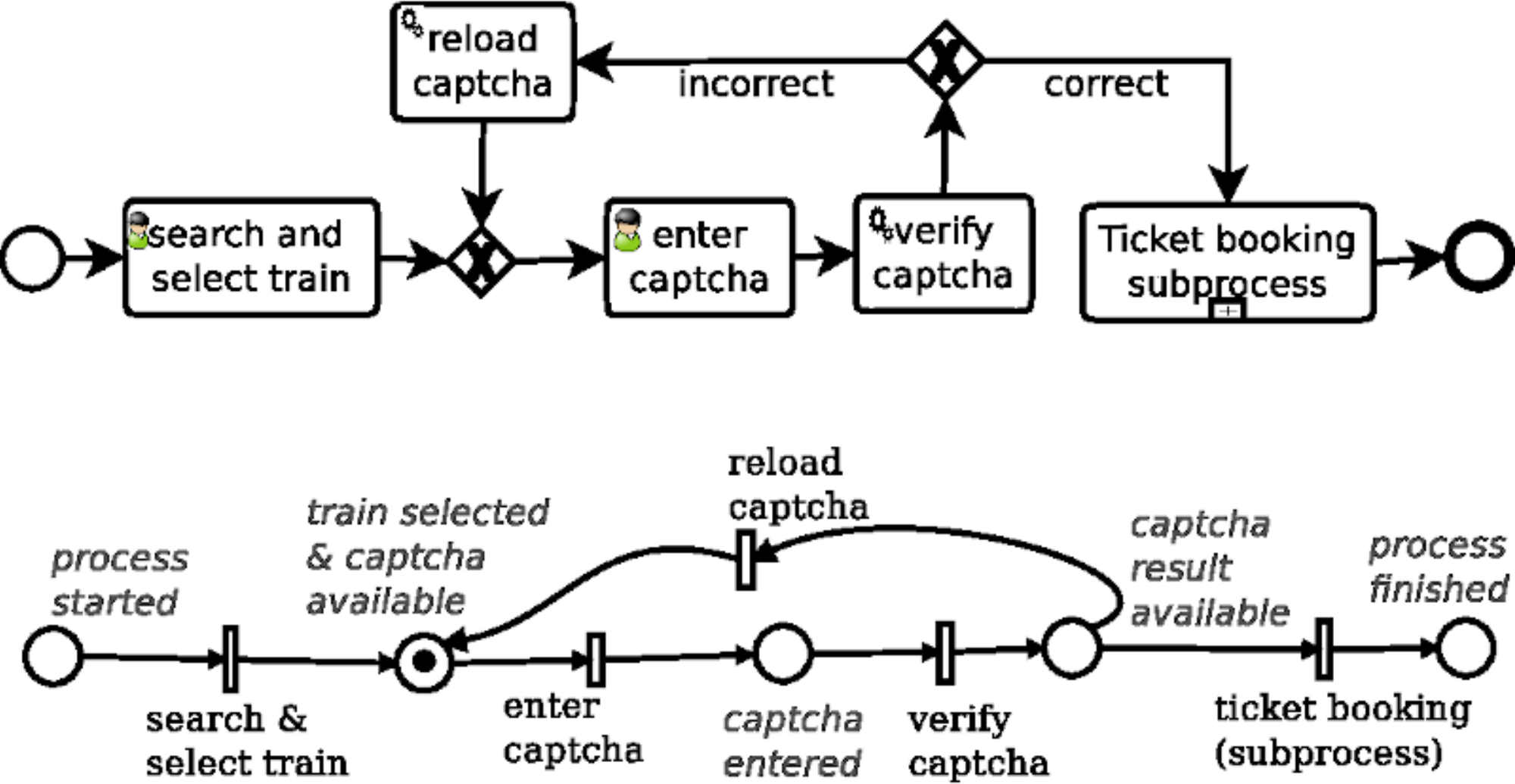}
  \caption{A Simple BPMN process and its WF-net}
    \label{fig:movie}
\end{figure*}

A class of Petri nets called WF-nets (Workflow nets) have been used in the 
paper to model workflows. 
First we informally discuss how the control-flow of a business process is 
modeled using Petri net with the help of an example as shown in Fig. 
\ref{fig:movie}. The figure depicts a BPMN model for railway ticket 
booking process, for which the corresponding WF-net is also given.
A WF-net consists of places (circles), transitions 
(bars or boxes), and arcs. Transitions are mapped to  activities/tasks/events, 
and
places model local conditions that serve as pre-/post-conditions for the tasks. 
This structure can model dynamics of discrete event-based systems  when {\em 
tokens} (dot) mark places. The set of all marked places make up a {\em 
marking} in the net, which explicitly represents the current state of the 
modeled system. 
A transition is {\em enabled} when all its 
{\em pre-places} are marked. An enabled 
transition 
{\em fires} by consuming one token from of its pre-places and by 
producing one token to each of its post-places, thereby changing the 
marking. A {\em firing sequence} or a {\em trace} is a sequence of transition 
firings from one marking to another. In Fig. \ref{fig:movie}, a token in
 place {\em train 
selected \& captcha available} describes one instance, 
in which the user has selected a train and the captcha is yet 
to be typed. The choice 
conditions {\em correct} and {\em incorrect} in the BPMN model is captured 
as a non-deterministic XOR fork in the WF-net control-flow architecture. This 
process 
model consists of only sequential and XOR fork/join control-flow constructs. 
Loops, AND fork/join etc. are examples of some other frequently occurring 
constructs in workflow modeling.

\subsection{\textbf{Existing Approaches to \textit{Change Region}}}
\label{relatedwork}
The foundational researches on change regions consider Petri net models of 
workflows. As noted earlier, a state in a workflow is modeled by a marking 
in the corresponding WF-net. Consequently, the \textit{same set of marked 
places} in two nets can be considered to be two equivalent states. The earlier 
approaches to change regions follow this notion of state-equivalence.

\subsubsection{\textbf{Computation of Change Region: Minimal-SESE Change 
Regions}}
The earliest notion of change region was proposed by Ellis et al. 
\cite{ellis1995dynamic}, though the computational aspect of it was not 
incorporated.
The change region for transfer validity was 
introduced by 
Van der Aalst \cite{van2001exterminating} on WF-net workflow model, defining 
the 
unsafe region for migration as the {\em dynamic change region}. In this 
approach, a change region is a connected 
subnet of the old net, which is the minimal 
Single-Entry-Single-Exit (SESE) region covering the structural changes.
The minimal-SESE change regions introduced in \cite{van2001exterminating}
were motivated by the argument that they can be abstracted as transitions in 
both the old and the new net covering the structural and behavioral 
difference so that the rest of the nets are structurally and behaviorally 
the same.
Sun et al. \cite{Sun2009284} present a variation of the change region 
generation algorithm given by Van der Aalst aiming at reducing the extent of 
the change region given in \cite{van2001exterminating}. 

\subsubsection{Adoption of Minimal-SESE Change Regions in the Literature}
Several other approaches use the minimal SESE change region approach.
Cicirelli et al. \cite{Cicirelli20101148} 
 adopted the minimal-SESE change regions for decentralized dynamic evolution, 
where the workflow is executed in a distributed environment. Hens et 
al. \cite{hens2014process} also adopt the approach of Van der Aalst to perform 
dynamic migration in each of the process fragments in a single location for 
distributed process evolution. The change region computation in their approach 
is performed on 
BPMN process models.
Gao et al. 
\cite{gao2013workflow} compute the SESE change region in SNS (special net 
structure) workflow models by step-wise abstraction of workflow fragments 
by a set of reduction rules \cite{gao2008process}. Zou et al. 
\cite{zouhybrid} have applied the SESE change region approach to workflow graph 
models, which are derived from the refined process tree of the workflow graphs 
\cite{vanhatalo2009refined}.
Sun et al. \cite{Sun2009284} point out that when it is required to find out a 
marking in the new net that has the same trace (firing sequence) as that of 
the given old marking, the minimal-SESE change region can be used to improve 
the efficiency to verify the trace equality between the markings.

\subsubsection{\textbf{The Problem of Overestimates in the Earlier Approach}}

\begin{figure*}[h!bt]
 \centering
 \includegraphics[scale=0.35]{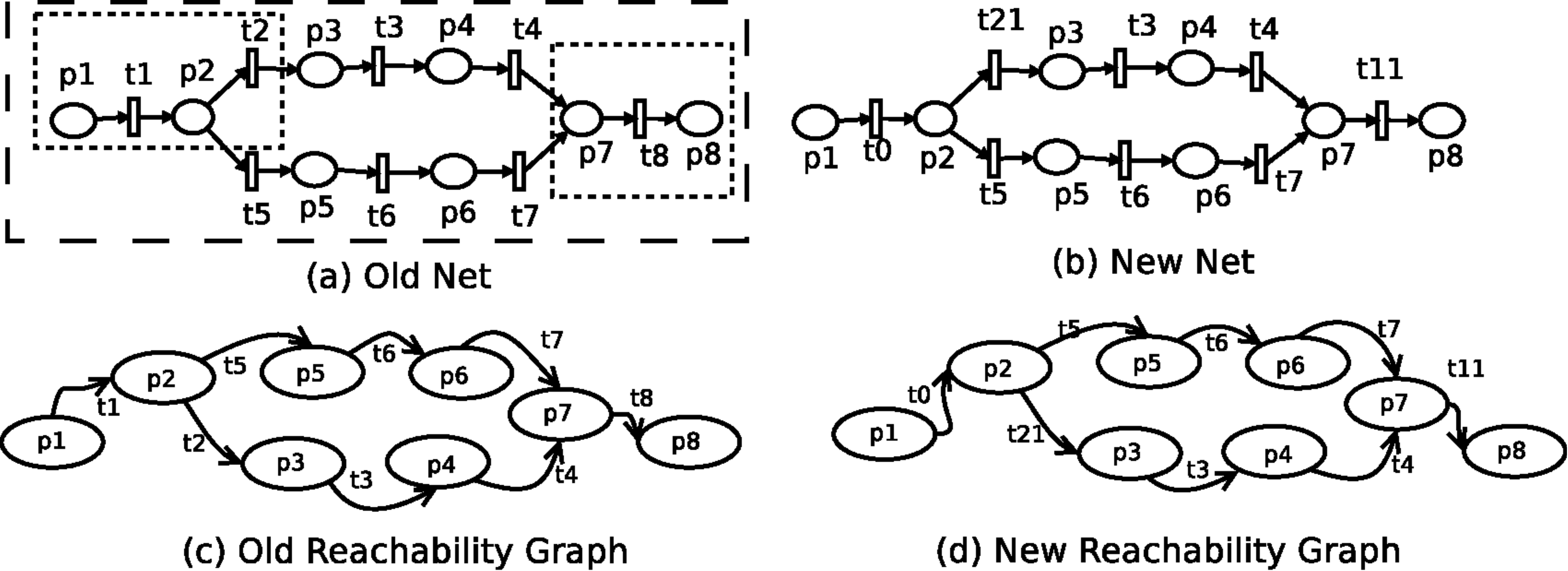}
 \caption{Overestimation in Minimal-SESE Change Region: for Transition 
Replacements}
\label{fig:andvdamore}
\end{figure*}

\begin{figure*}[h!bt]
 \centering
 \includegraphics[scale=0.35]{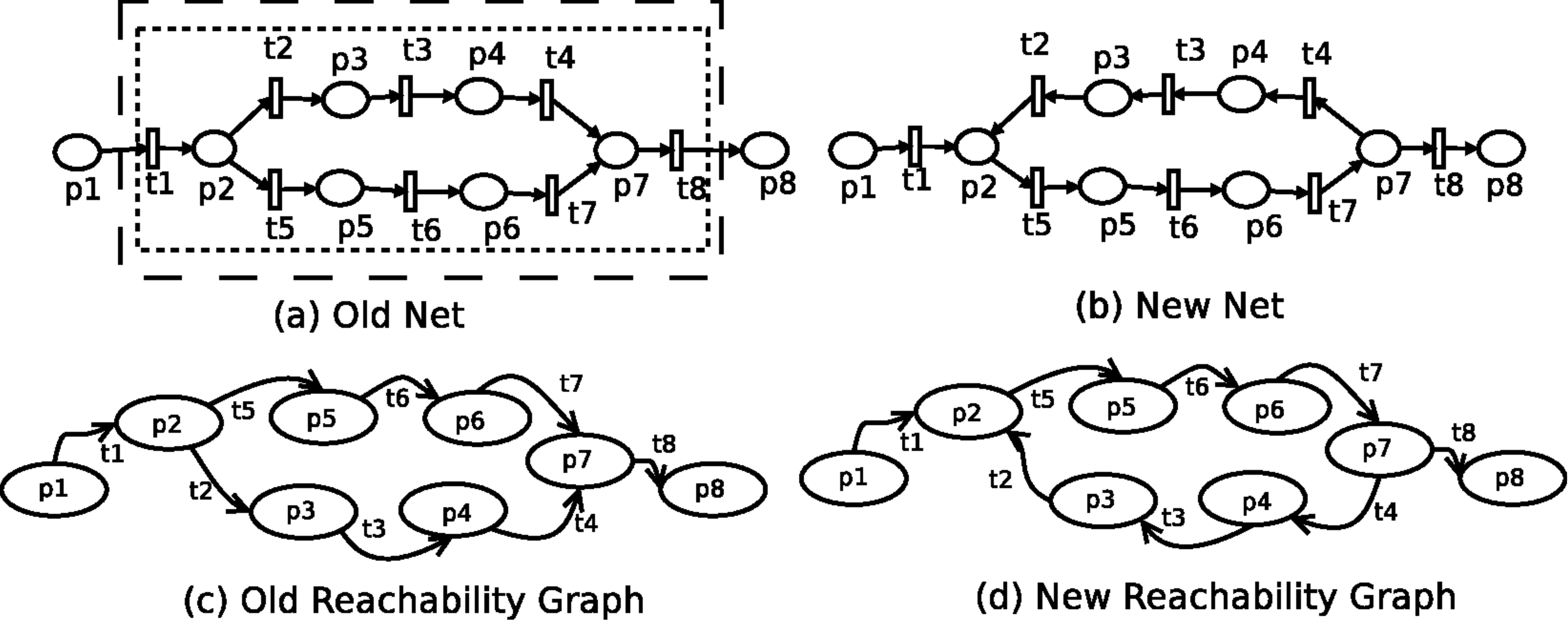}
 \caption{Overestimation in Minimal-SESE Change Region: for change 
from XOR to Loop}
\label{fig:loopxorcr}
\end{figure*}

In the minimal-SESE 
approach, a change region is composed of places, transition and arcs.
The change region generation algorithm as given by Van der Aalst in 
\cite{van2001exterminating} gives an overestimation of non-migratability as 
pointed out in the same paper. The example of this case is  reviewed in Fig. 
\ref{fig:andvdamore}. In the figure, net (a) is the old net. Transitions $t_1$, 
$t_2$ and $t_{8}$ are replaced by $t_0$, $t_{21}$ and $t_{11}$ respectively 
giving the new net as shown in net (b). The corresponding reachability graphs 
are depicted in Figs. \ref{fig:andvdamore}(c)-(d). It can be observed that, 
though the transitions among states are different in the new state-space as 
compared to the old, the {\em states} or the markings in both the graphs are 
the same. In other words, every old state is a valid state in the new net. 
Hence, every marking in the old net is migratable.

The approach starts with computing the
 \textit{static change region}, which is  composed of all the net elements 
(places and 
transitions) involved in the arcs that are present in only one of the nets but 
not in both. Arcs $(p_1, t_1)$, $(t_1, p_2)$, $(p_2, t_2)$, $(p_7, t_8)$, and 
$(t_8, p_8)$ in net (a) are such arcs. The resultant static change regions 
cover 
two subnets is shown by dotted boundaries in the figure. It consists of net 
elements $p_1$, $t_1$, $p_2$, $t_2$, $p_7$, $t_8$, and $p_8$ from the old net. 
Since the structural difference is not sufficient to isolate all the behavioral 
differences, a \textit{dynamic change region} is defined as the minimal 
SESE-region 
(Single Entry Single Exit region) covering a static change region. Such a 
SESE-region ensures exactly the same state-space outside the change region.
 For example, for the above nets in Fig. \ref{fig:andvdamore}, the dynamic 
change region (shown as dashed boxes) additionally includes places $p_3$, 
$p_4$, 
$p_5$, and $p_6$, besides the places $p_1$, 
$p_2$, $p_7$ and $p_8$ in the static change region. However, the above type of 
change regions may include overestimates since structural changes may not 
result in changes in the reachable markings, which is shown in Fig. 
\ref{fig:loopxorcr}.

Moreover, in their approach, an improved minimal change region is obtained
by discarding the boundary places. For this example, it excludes $p_1$ and 
$p_8$ 
from the dynamic change region.
However, even after discarding the boundary
places, the cases of false negatives are not completely eliminated.
An example of overestimation in improved change regions is given in Fig. 
\ref{fig:loopxorcr}, in which, 
  an XOR-block is changed into a loop-block. All
the markings in the old net are migratable based on validity though the
behavior of token flow is changed. Hence the entire change region is an 
overestimation.

The algorithm for \textit{change node generation} as given by Sun et al. 
\cite{Sun2009284} also suffers from overestimation since the change region that 
it generates is same. Cicirelli et al. \cite{Cicirelli20101148} later point out 
that the algorithm in \cite{Sun2009284} does not improve the change region over 
the earlier minimal-SESE approach.

\subsection{\textbf{Our Approach}}
\begin{figure*}[h!bt]
\centering

 \includegraphics[scale=0.3]{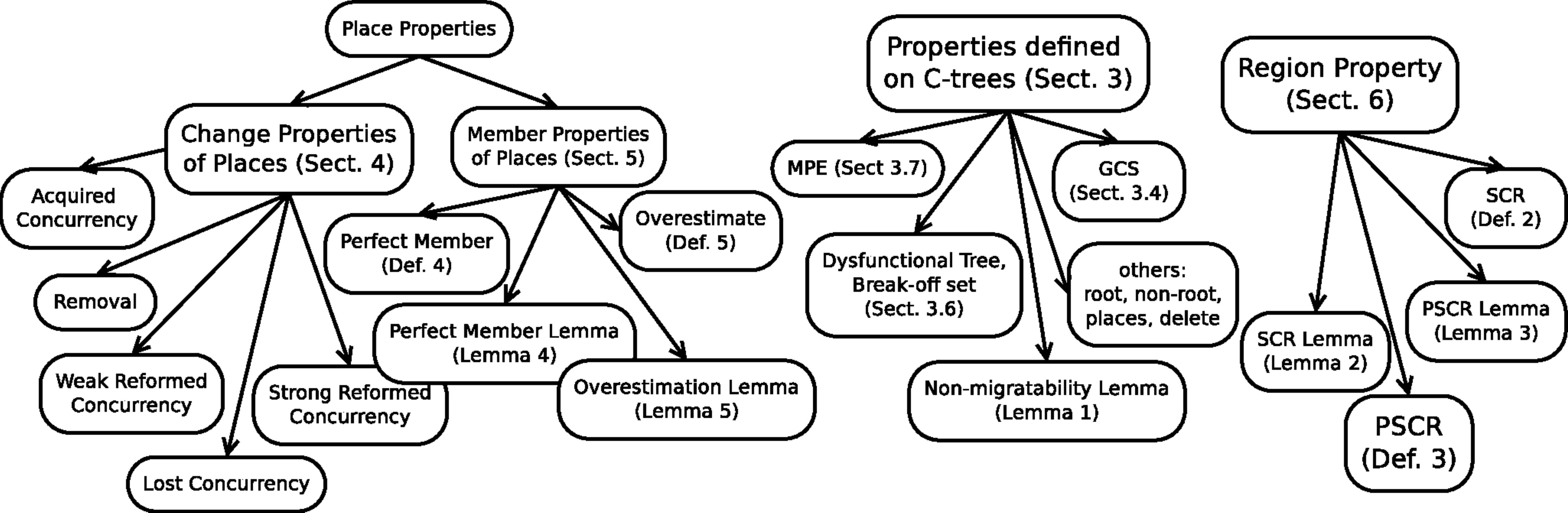}
  \caption{Contributions}
    \label{fig:conceptTree}
\end{figure*}

Our approach to change region focuses on how the markings in the old net are 
changing, which leads to investigation of concurrency conflicts. Keeping 
non-migratability w.r.t. transfer validity at the core,  the paper 
characterizes 
various properties at place level (\textit{place properties}) and at net/subnet 
level ({\em region properties}) and then constructs Lemmas based on them to 
result in computation of the two kinds of change regions. The 
contributions are identified in Fig. 
\ref{fig:conceptTree}.

\subsubsection{\textbf{Place Properties and Region Properties}} As depicted in 
the figure, place properties are of two kinds: \textit{Change properties} and 
\textit{Member properties}. \textit{Change  properties}  capture the change in 
concurrency in the net due to  a given place. Concurrency may change due to 
reasons such as removal of the place or loss or acquisition of concurrency.  

\textit{Member properties} indicate the contribution of the place to change 
region, which can  either be as an \textit{overestimate} causing 
false-negatives, or as a \textit{perfect member} fully contributing to 
non-migratability. An overestimated place is involved in both migratable and 
non-migratable markings. A place that is a perfect member is always involved in 
only non-migratable markings.

{\em Region properties} are about nets or subnets and not about individual 
places. The definitions of SCR and PSCR fall under this category. 

Using the above place properties and region properties,  lemmas are to show   
(i) constructibility of SCR in terms  of overestimation and perfect members 
(Lemma 2), (ii) necessary and sufficient condition of existence of PSCR (Lemma 
3), (iii) necessary and sufficient condition for a place to be a perfect member 
(Lemma 4), and (iv) necessary and sufficient condition for a place to be an 
overestimation (Lemma 5). Lemma 2, 3, 4 and 5 are used in  
computation of PSCR and SCR. 

\subsubsection{\textbf{The C-tree Structure}}
 
We use {\em C-tree} structures and the notion of Generator of 
Concurrent Submarking (GCS) introduced earlier by us in \cite{DBLP:conf/apn/PradhanJ16}.  C-trees are built out of nets by capturing 
only the concurrency aspect of the nets and by removing the rest of the 
information in the net.  Given a place, GCS extracts its peer concurrent part in the C-tree. 

The following new properties are developed on C-trees to provide computational means to assert place 
properties concerning migratability. Marking Preserving Embedding (MPE) between two C-trees ensures full-migratability 
between the corresponding nets. Dysfunctional tree and Break-off 
set assist in identifying perfect members. 

Consequently, the C-tree structure has been used as a basic 
structure in  computations that make region sets out of these places. 
For a 
given a pair of nets in terms of C-tree structures, the condition for existence 
of non-migratability is provided by Lemma 1.
The lemma thus gives the pre-condition for computing the change regions.

\subsection{Organization}
The rest of the paper is organized as follows.  
Section \ref{prel} describes the background on  WF-nets, defines the scope 
of block-structured WF-nets used in the paper, the notion of consistency, and 
the use of change region in dynamic migration. 
Section \ref{background} develops the C-tree structure and the properties 
defined on it for use in the rest of the paper. 
Section \ref{propsection} develops the change properties of 
places. Section \ref{memprop} discusses the member properties and the related 
lemmas. Section \ref{minimality} develops the region properties and the related 
lemmas, discusses computation of SCR and PSCR. It also brings out the utility of 
the approach showing improvement over the earlier approach of minimal-SESE 
regions through an example case.
\begin{figure*}[h]
\begin{tabular}{c c}
\hspace*{-1.2cm}{\it
 \begin{tabular}{p{5mm} l c l p{5mm} p{5mm} l c l p{5mm} }

&Net &$\rightarrow$ &Pnet&& &&&&\\
&Pnet	&$\rightarrow$ &{\bf Place} &&&&&&\\				
&	&&$|$ Pnet {\bf Trans} {\bf Place} 	&&&&&&\\	
&	&&$|$ Pnet {\bf Trans} loop {\bf Trans} Pnet&&&&&&\\
&	&&$|$ Pnet {\bf Trans} and {\bf Trans} Pnet	&&&&&&\\
&	&&$|$ Pnet xor Pnet 	&&&&&	&\\
\end{tabular}
}
&\hspace*{-4cm}
{\it
 \begin{tabular}{p{5mm} l c l p{5mm} p{5mm} l c l p{5mm} }
&Tnet	&$\rightarrow$&	{\bf Trans}	$|$ {\bf Trans} Pnet {\bf 
Trans}&&&&&&\\ 
&loop	&$\rightarrow$ &\textbf{\{} Pnet 
\textbf{\} \{} Tnet \textbf{\}}&&&&&  &\\

&xor&$\rightarrow$ &\textbf{[} Tnet\textbf{ ] [} Tnet \textbf{]}&&&&&&\\	
&&&$|$ \textbf{[} Tnet \textbf{]} xor&&&&&&\\
&and	&$\rightarrow$ &\textbf{( }Pnet \textbf{) ( }Pnet \textbf{)}&&&&&&\\
&&&$|$ \textbf{(} Pnet \textbf{)} and&&&&&&\\
\end{tabular}
}
\end{tabular}\caption{ECWS Grammar}
\label{fig:grammar}
\end{figure*}

\section{Background}\label{prel}

This section provides the background of Workflow nets, marking reachability, 
consistency and change region. 

\subsection{WF-nets}
This section briefly reviews the formal definition of Workflow nets 
(WF-nets) as founded by Van der Aalst \cite{van1998application}.

The structure of a WF-net $N$ is defined as a tuple $(P,T,F)$, where $P$ is a 
finite set of places, $T$ is 
a finite set of transitions, and $F\subseteq (P\times T)\cup (T \times P)$ is a 
finite set of arcs. 
The properties of WF-nets is reviewed below:

\begin{itemize}
 \item \textit{Unique Source:} $init\in P$ is the only source place. The source 
place is identified by condition $\bullet init = 
\emptyset$. 
\item {\em Unique Sink:} $end\in P$ is the only sink place. The sink place is 
identified by condition $end\bullet  = \emptyset$ 
\item
{\em Connectedness:}
Apart from the source place $init$ and the sink place $end$, all other places 
and all transitions appear on at least one directed path from 
$init$ to $end$.
\item {\em Unique Initial Marking:} One token in the source place $init$ gives 
the 
initial marking $M_0$. 
\item {\em Unique Terminal Marking:} When the token reaches  the sink place,  
all other places $p \in P\backslash\{end\}$ are in  unmarked state.
\item {\em Behavioral Well-formedness:} (i) No Dead Transition:  Every 
transition $t\in T$ can be fired from some marking $M$  reachable from $M_0$  
(ii) Proper Termination:
 Every marking $M$ 
reachable from $M_0$ through a firing sequence
eventually reaches terminal marking.

Set $\mathbb{R}(M)$  gives the set of markings reachable 
from a given marking $M$.
Therefore, $\mathbb{R}(M_0)$ is the valid set of markings in the corresponding 
WF-net.
\end{itemize}

WF-nets are {\em safe} nets, i.e. no marking that is reachable from the initial 
marking marks a place in the net with more than one token. Consequently, in 
every possible 
marking in the net-dynamics, a place can hold only one token at most. 
Mathematically, a marking is often represented as a column matrix, where the 
number of rows is equal to the total number of places in the given net; the 
elements corresponding to the marked places are $1$ and the elements 
corresponding to the unmarked places are zero. For example, the marking shown 
in Fig. \ref{fig:movie} is $(0 1 0 0 0)^T$. However, in this paper, we stick to 
the traditional definition of marking in a net, which describes 
\textit{marking as a set of marked places}. It is a more generic mathematical 
form than implementation-friendly matrix-form. For the marked net in Fig. 
\ref{fig:movie}, the marking is {\em
\{train selected \& captcha available\}}. 

\subsection{Block-Structured WF-nets}
The theory  developed in this paper are applicable to all 
structured workflow process models composed of sequence flows, choices, 
concurrent branching, and loops.
Block-structured nets have been noted to cover a large portion of 
the 
business process logics \cite{van1998application}.
In our analysis on migratability, the old and the new workflows 
modeled as WF-nets are assumed to be  
well-formed, i.e. they are assumed to be constructible through a 
structured composition grammar. 
For our purpose,  the Extended Compact Workflow 
Specification (ECWS) grammar \cite{DBLP:conf/apn/PradhanJ16} as enlisted in 
Fig. \ref{fig:grammar} is followed. 
 ECWS uses   labels for places and transitions, and 
bracketed enclosures as block delimiters to express the logic of 
concurrency, exclusive-choice and loops.

\begin{figure*}[h!bt]
\centering
 \begin{tabular}{ p{4.7cm} p{8cm} }
 
 \includegraphics[scale=0.32]{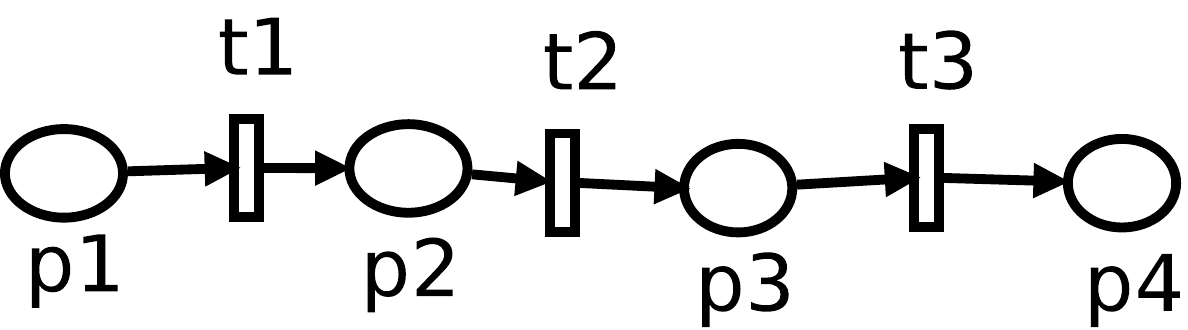}
 
 {\bf (a)} $p_1$$t_1$$p_2$$t_2$$p_3t_3p_4$& 

\includegraphics[scale=0.32]{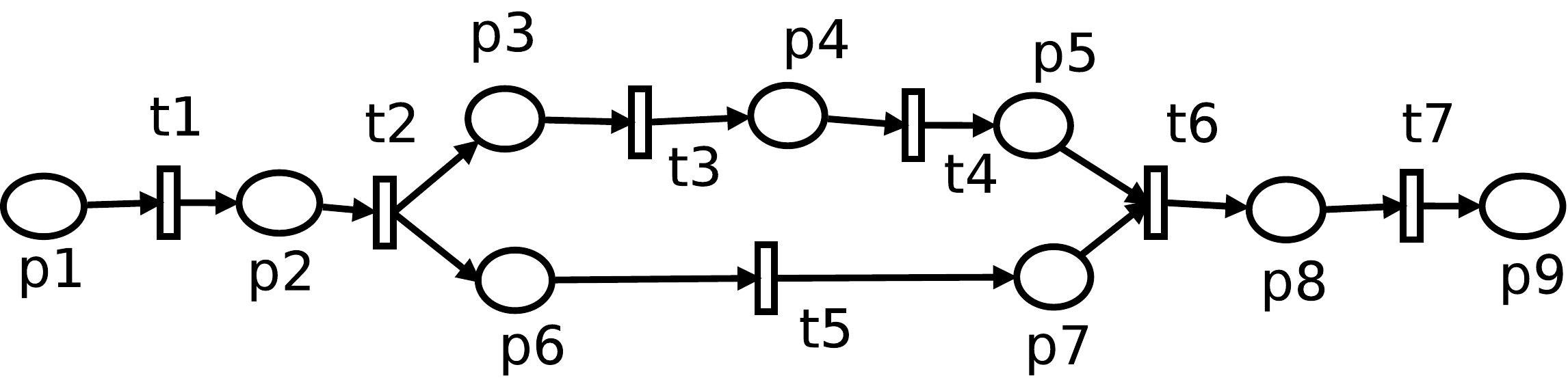}

{\bf 
(b)} 
 $p_1t_1p_2t_2($ 
$p_3$$t_3$$p_4$$t_4$$p_5$$)$$($$p_6$$t_5$$p_7$$)$$t_6$$
p _8$$t_7$$p_9$\\\\

\includegraphics[scale=0.32]{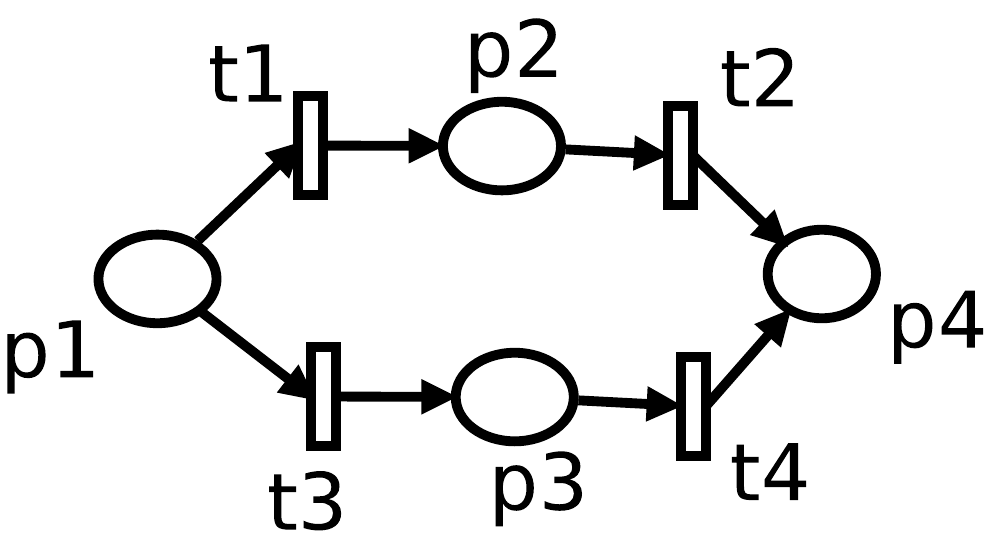}
 
{\bf (c)} $p_1[$$t_1$$p_2$$t_2$$]$$[$$t_3$$p_3$$t_4$$]$$p_4$&

\includegraphics[scale=0.32]{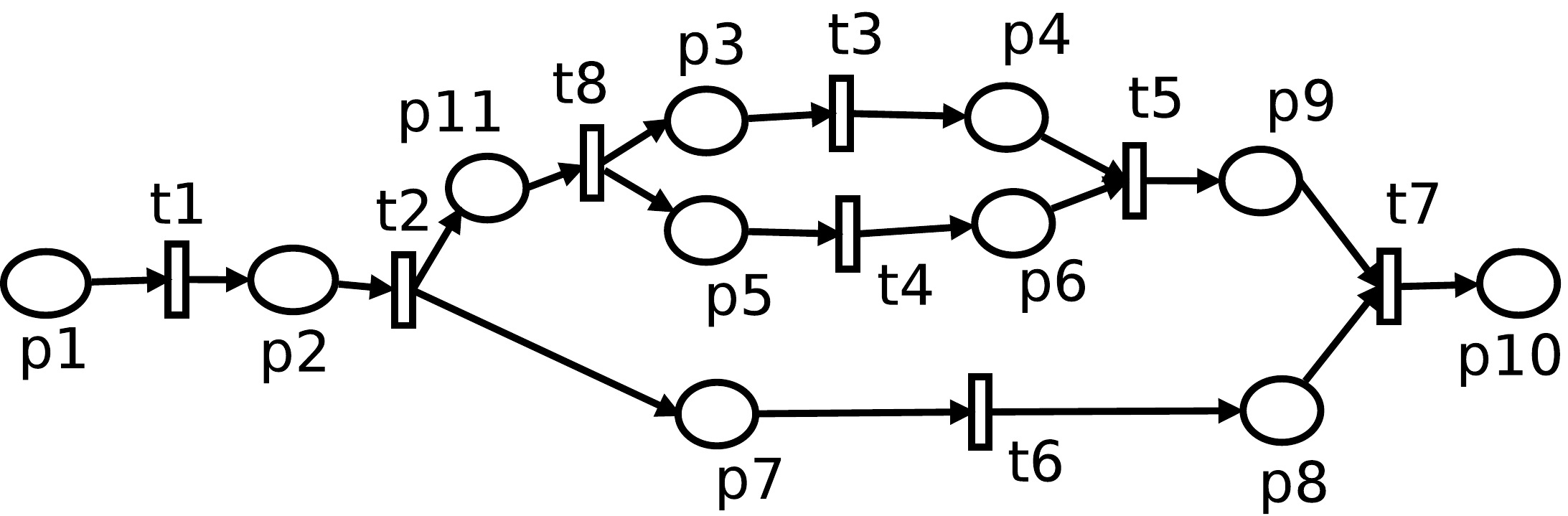}

{\bf 
(d)} 
 $p_1$$t_1$$p_2$$t_2$$(p_{
11} t_8 
$$($$p_3$$t_3$$p_4$$)$$($$p_5$$t_4$$p_6$$)$$t_5$$p_9$$)($$p_7$$t_6$$p_8$)$t_7p_{10}$ \\\\
\includegraphics[scale=0.32]{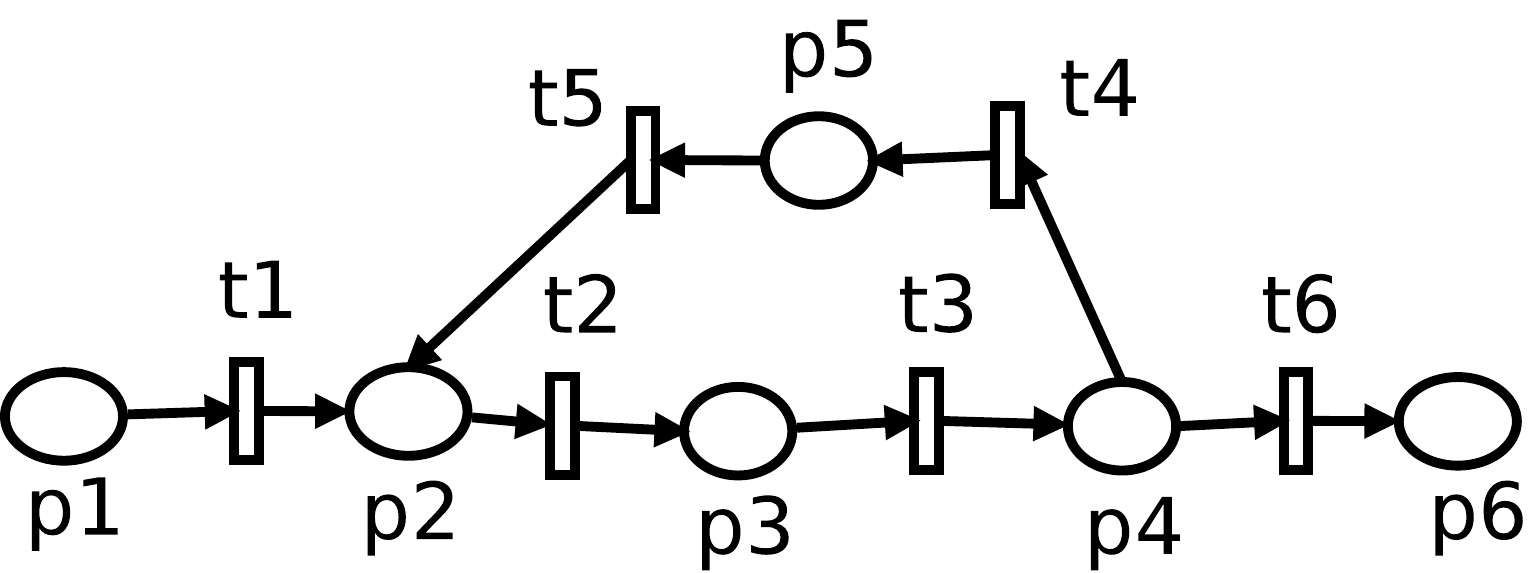}

{\bf (e)} $p_1t_1\{p_2t_2p_3t_3p_4\}\{t_4p_5t_5\}t_6p_6$
&

\includegraphics[scale=0.32]{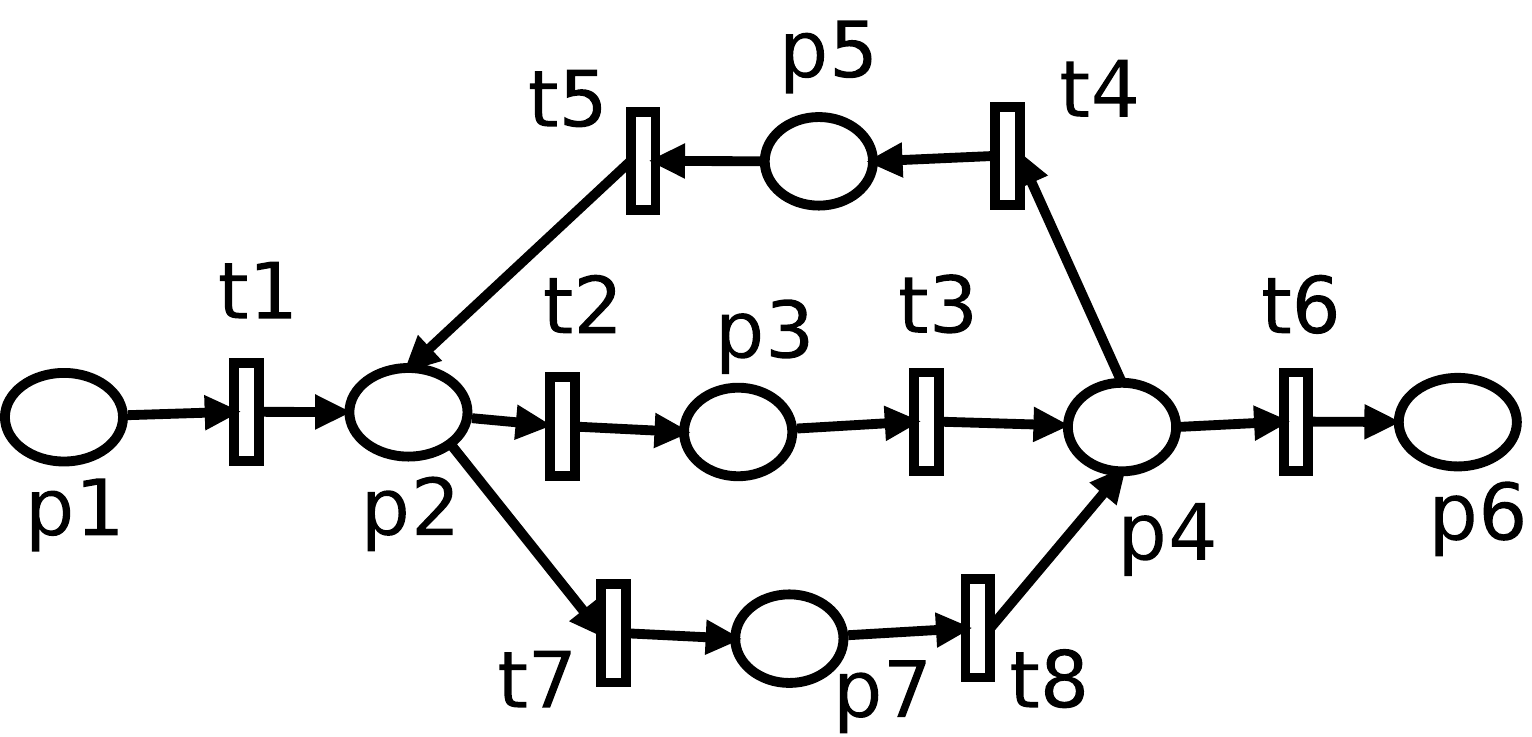}

{\bf (f)} $p_1t_1\{p_2[t_2p_3t_3][t_7p_7t_8]p_4\}\{t_4p_5t_5\}t_6p_6$

 \end{tabular}
  \caption{ECWS Specification Examples}
\label{fig:notation}
\end{figure*}

Fig. \ref{fig:notation} presents examples of workflow models specified in ECWS. 
Figs. \ref{fig:notation} (a)-(f) respectively show a (a) simple sequence, (b) 
parallel fork-join block inside
a sequence, (c) simple choice-block, (d) nested parallel fork-join block with 
an 
inner parallel fork-join, (e) a loop within a sequence, and lastly (f) a loop 
which has a choice-block inside its forward branch. The respective ECWS 
specifications
can also be found in the figure.

\subsection{Marking Reachability}

Given a WF-net, its reachability graph provides the markings (states) that are 
reachable from its initial marking, and the corresponding state-transitions. 
Due to presence of concurrency (AND-blocks), number of states in the 
reachability graphs is usually much larger than the number of places in the 
net. In absence of concurrency, individual places correspond to states.
Figs. \ref{fig:andvdamore} and \ref{fig:loopxorcr} show reachability graphs 
corresponding to the given WF-nets.

\subsection{Consistency Criterion}
\label{overestimation}

Migratability of a state (token-marking) depends on the consistency
 criterion  for migration, i.e. the notion of {\em migration equivalence} 
between two states. Variants of consistency notions have been identified in the 
literature. These can be  classified into trace-based and state-based models.  Rinderle et al. \cite{rinderle2004correctness} discuss consistency notions adopted by earlier approaches, 
out of which, most commonly used notions rely on matching the 
{\em traces} between the equivalent states. Intuitively, the same set of completed 
activities before and after the migration ensures the correctness of migration, 
which in turn leads to successful execution of the workflows after migration. 
Ellis et al. \cite{ellis1995dynamic} present the earliest notion of consistency in terms of \textit{trace equality}. Variants of trace-based consistency criteria were defined in another work of Rinderle et al. \cite{rinderle2008relaxed}. In our previous work \cite{Pradhan:2014:TTP:2590748.2590765} 
\cite{mecatalog15} on trace-based consistency in migration, we devised a catalog-based approach for  
migration of block-structured WF-nets.
Our work on lookahead consistency 
\cite{lookahead} defines a class of future-trace based consistency. 

On the other hand, a state-based consistency criterion called \textit{ transfer validity} was  
formulated by Van der Aalst in \cite{van2001exterminating} in the 
context of WF-nets \cite{van1998application}. This criteria was adopted notably  by 
Van der Aalst and Basten \cite{van2002inheritance} and by Cicirelli et al. \cite{Cicirelli20101148} 
for WF-nets. It was also applied by Hens et al.  \cite{hens2014process} for BPMN \cite{bpmn2.0} models of workflows. 
This paper uses the notion of {\em 
transfer validity} that establishes equivalence based on the {\em present 
attributes} of the states rather than the past or future. 
The notion of {\em transfer validity} is revisited below. We also highlight its motivation and applicability with the help of an example.

\begin{defi}\label{def:consistency}
\textbf{Consistency (Transfer Validity)}: A marking $M$ in a WF-net $N$ is 
consistent with a marking 
$M'$ in 
WF-net $N'$ iff $M = M'$,  $M \in \mathbb{R}(M_0)$, $M' \in 
\mathbb{R}(M_0')$,  where $M_0$ and $M_0'$ are initial markings of $N$ and 
$N'$ respectively. A marking in a net is a set of places holding tokens in that 
net. 
The places are identified by their labels.
\end{defi}

Using the definition, we obtain set $\mathbb{R}(M_0)$ $-$ $\mathbb{R}(M_0')$ as 
the 
collection of non-migratable markings in $N$, since these markings do not 
satisfy the 
consistency criterion. 
In general,  the applicable scenarios are the 
situations where strict equivalence notions such as the same trace of completed 
activities are not required, and marking equality indeed captures equivalent 
states in terms of business semantics. Therefore, the applicability depends on the required business goal. In 
the following, an example is provided to illustrate such a case.
\begin{figure*}[h!bt]
\centering
 \includegraphics[scale=0.34]{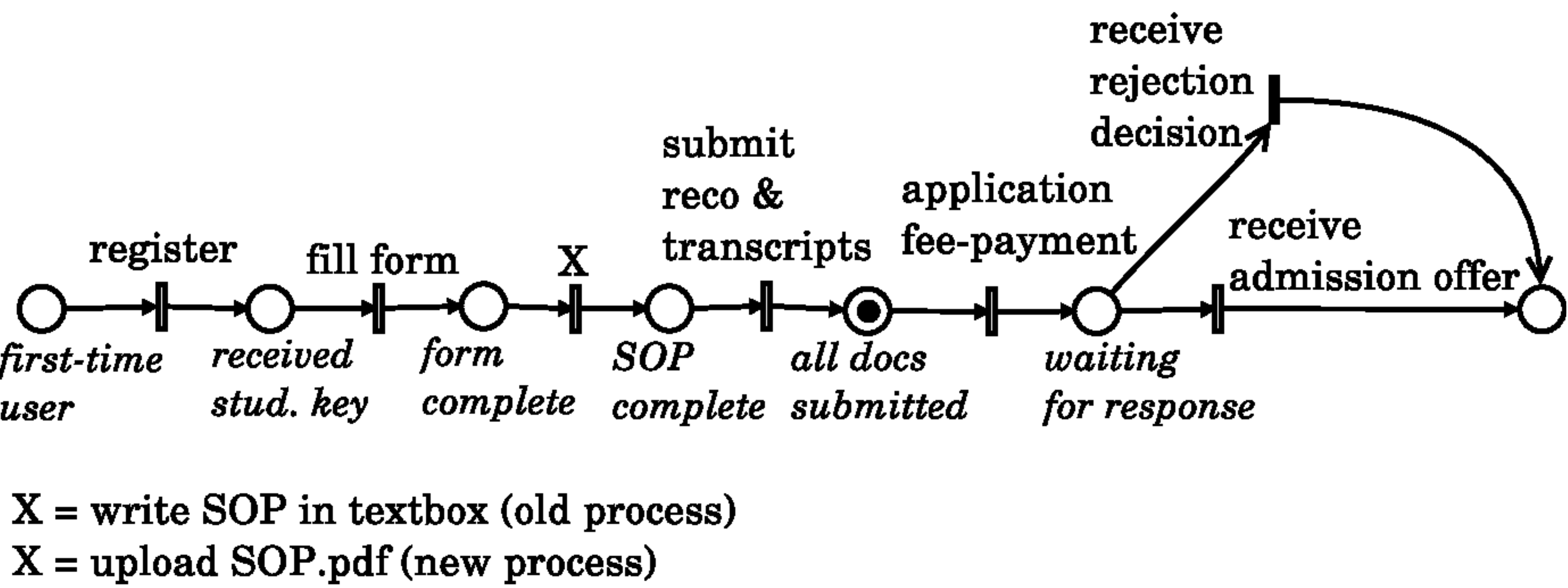}
  \caption{Example of Migration case with Consistency Criterion of Transfer 
Validity}
    \label{fig:uniapp}
\end{figure*}

Fig. \ref{fig:uniapp} shows a semester-long workflow for an online application
process in a university. The workflow starts with registration of an applicant 
through admission portal
website, which generates a registration ID. Then 
an application form is provided to be filled in. Next, a text-box  is 
provided to type out
the statement of purpose (SOP).  The rest of the workflow consists of 
submission of necessary documents, payment and receipt of admission decision. 
After running the workflow for a month, the admission committee decides to 
change the application task of SOP submission through typing out in text-box by 
replacing it with an upload-pdf activity. The rest of the workflow  remains 
unchanged as shown in the figure.
Though the application task for SOP submission is different in the two 
processes, they both achieve 
the same business goal. Consequently, their post-place has the same label 
``SOP-complete'' in both the nets, and once a token reaches this place, it does 
not matter actually how SOP was submitted.
As a result, an 
applicant, who
has written the SOP in the text-box and now waiting for response when the new 
process is launched, dynamic migration allows her application to be migrated 
to the equal marking in the new process. The marking shown in the figure is one 
such state, which is consistent in both of the
workflows as per Def. \ref{def:consistency}. Conclusively, depending on the 
business 
situation, trace equality may not be necessary to establish migration 
equivalence, and transfer validity of marking may be sufficient to achieve 
correct migration.

\subsection{An Illustration of Change Regions}

\begin{figure*}[h!bt]
\centering
\includegraphics[scale=0.32]{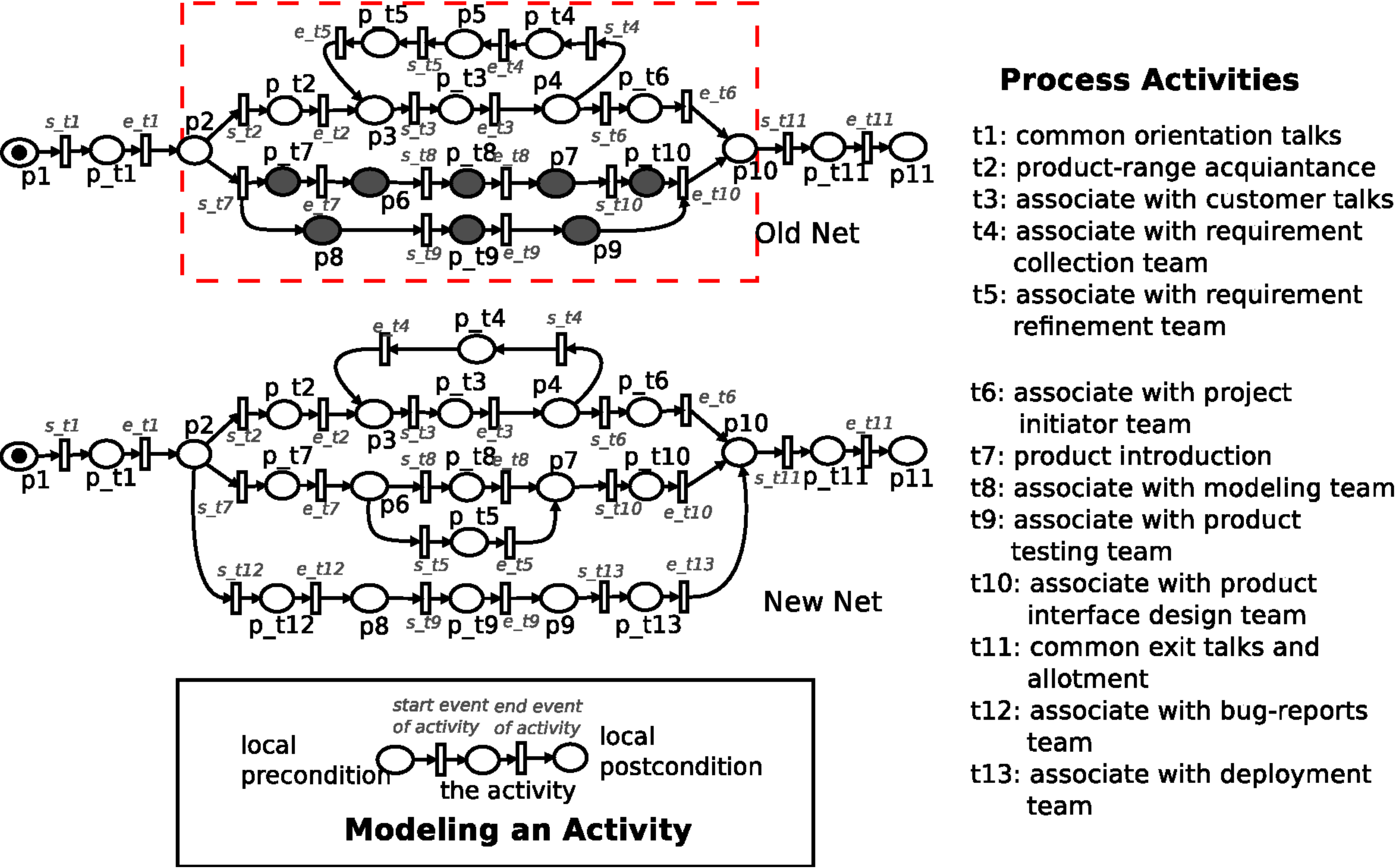}
\caption{A Training Process of a Company}
\label{fig:cr}
\end{figure*}

Now we describe the notion of change region in dynamic migration through a 
practical example. Fig. \ref{fig:cr} depicts a training 
process in a company for its continuous inflow of 
recruits. 

\subsubsection{The Old Process}
In the old process, two alternatives for the trainees are available 
after a common orientation program. The top path is for marketing trainees, and 
the bottom path is taken by the design trainees. Marketing trainees are exposed 
to the product-ranges, requirements, customer interactions etc through a 
well-designed loop till completion. The design trainees are exposed to 
products, 
testing, modeling and interface design tasks. The individual tasks have been 
identified in the figure. The reachability graph for this net has 29 states.

\subsubsection{The New Process}

The training department makes some changes to the process
based on a feedback that design trainees need to understand some
portion of requirements, and since the division of testing got
separated from product design, an extra option be created for testing
trainees. The process is therefore evolved as shown in the new net (with its 
initial marking) in the figure. The training department shifts one task 
($p_{t5}$) from the 
upper part
into the lower part, and it splits the lower AND composition into
an XOR composition creating a new path with additional tasks in it. The 
corresponding net has 22 states in the reachability graph.

\subsubsection{Use of Change Region in Migration}
This being a 3-months training program, the management explores whether
the ongoing trainees can be directly migrated into the new process. 
Mere enumeration of markings of both the nets result in $29\times22$ ordered 
pairs, out of which, the identity relations are the migratable markings. In 
general, business processes involve concurrency due to which the state-space 
is usually large, and therefore, searching through the state-space is 
computationally costly.
Use of change region does not require finding out such marking mappings. 
Instead, a region in the old net is detected as unsafe (maximum size of which 
is the number of places in the net, 22 in this case) in the change region 
approach. If a given marking overlaps with the region, it is non-migratable. In 
Fig. \ref{fig:cr}, the red-dotted box cover the places computed by the 
minimal-SESE approach. The shaded places identify the places in the change 
region by our PSCR approach.
A trainee continues in the old process till the marking is in the change region 
and migrates as soon as the marking comes out of it.
Using the PSCR, when a token arrives in 
any member place in set $\{p_{t7}, p_6, p_{t8}, p_7, p_{t10}, p_8, p_{t9}, 
p_9\}$ (places shaded in the old net), 
the marking is non-migratable.

\section{Foundations: C-trees and their Properties}\label{background}

This section develops a structure called the C-tree, which is a structural 
model for capturing concurrency structures underlying all markings in a given 
ECWS-net. By using C-trees, we analyze the marking-based conflicts for 
consistent migration as per the consistency criterion given in Def. 1, without 
going into the state-space.

\subsection{\textbf{Marking Generator Sets (MGS)}} Our structure-based 
approach develops based on a notion of 
{\em Marking Generator Sets}, which is a nested set of {\em 
places} computed for the entire net. It represents concurrent sets of places 
organized hierarchically as per 
the net structure, which is a structural abstraction of all possible markings 
of a given ECWS-net.  It can be noted that, the MGS can be 
easily obtained from the ECWS specification by Algorithm \ref{algo:ecws2mgs}.
The depth of the MGS structure gives the 
depth of concurrent nesting in the net. If there is no concurrency, MGS has no 
nesting.  

\begin{figure*}[h!bt]
 \centering
 \includegraphics[scale=0.35]{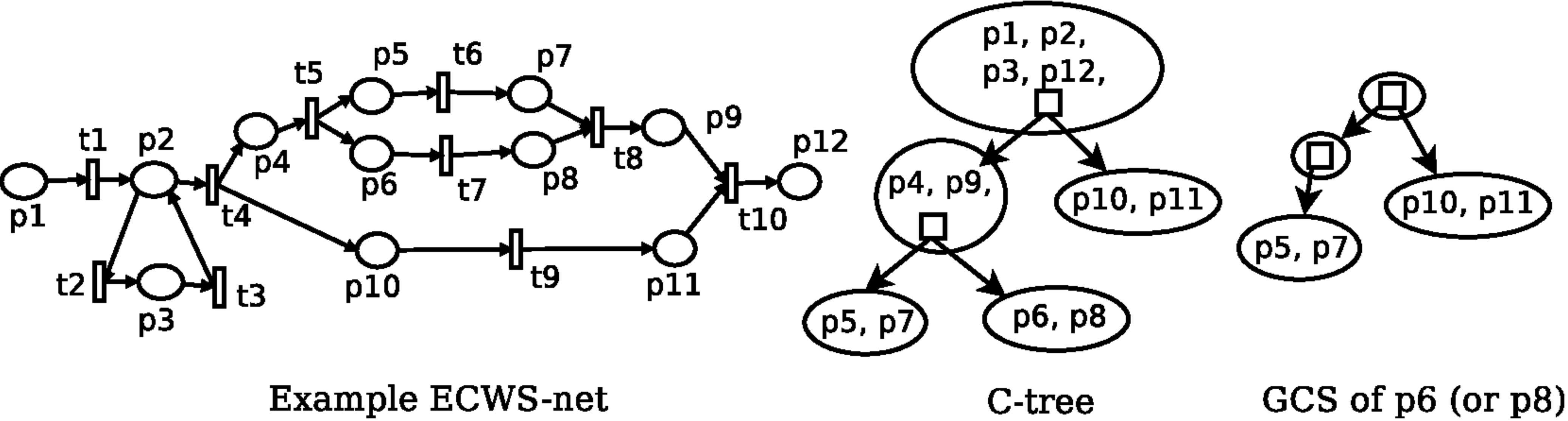}
 \caption{C-tree and GCS Examples}
 \label{fig:ctree}
\end{figure*}  

{\color{red}
\setlength{\intextsep}{5pt}
\begin{algorithm}
{
constructMGS(terminal/non-terminal $e$)

\If{$e$ is a {\em place}}
{
  \Return label($e$)\;
}
\ElseIf{$e$ is AND non-terminal}
{
    tuple $t$ $\leftarrow$ $\emptyset$\;
    $ex$ $\leftarrow$ set of terminals/terminals in grammar expansion of $e$\;
    \For{ each $exp$ in $ex$}
    {
        $t$.add(constructMGS($exp$)\;
    }
    \Return $t$
}
\ElseIf{$e$ is non-AND non-terminal}
{
    set $s$ $\leftarrow$ $\emptyset$\;
    $ex$ $\leftarrow$ set of terminals/terminals in grammar expansion of $e$\;
    \For{ each $exp$ in $ex$}
    {
        $s$ $\leftarrow$ $s$ $\cup$ constructMGS($exp$)\;
    }
    \Return $s$ 
}
}
 \caption{Generating MGS}
 \label{algo:ecws2mgs}
\end{algorithm}
}

\begin{defi}\label{defi:mgs}
{\bf Marking Generator Set:}
MGS is created as follows:
(i) A set is created for collecting all places in sequences,
XOR-blocks, loop-blocks in same level in the block structure. (ii) A tuple is created for every concurrent block.  Elements in the tuple can be constructed by either (i) or (ii). Therefore,
MGS can be recursively defined as:
{\bf MGS} =  set mutually exclusive {\bf element}s, where
{\bf element}  = {\bf place} or {\bf tuple} of concurrent {\bf element}s.
\end{defi}

 For example, for the net in Figure \ref{fig:ctree}, the MGS is 
\{$p_1$,$p_2$,$p_3$,(\{ $p_4$,(\{$p_5$,$p_7$\},\{$p_6$,$p_8$\}), $p_9$\},\{$p_{10}$,$p_{11}$\}),$p_{12}$\}.

\subsection{\textbf{Obtaining MGS from ECWS specification}} A marking generator 
set is derived from the ECWS-net by recursively starting 
from the top level net as follows:
(i) A top-level set is created for collecting all places in sequences, 
XOR-blocks, loop-blocks in top-level.
 (ii) An internal set is created for every inner concurrent block.
(iii) In every concurrent set, a top-level set is created for each concurrent 
branch, again to collect all places in sequences, XOR-blocks, loop-blocks in 
that branch.

Presence of loop in a net does not increase the nesting level of the member sets. Places in Loop, XOR and sequence are treated similarily in MGS. The net depicted in Figure \ref{fig:ctree} contains a loop that starts and ends at $p_2$. In the above MGS representation places $p_2$ and $p_3$ appear at the outer most level of set nesting.

\subsection{\textbf{Conjoint Tree representation of MGS}}
\textit{Conjoint tree} (C-tree) is the tree structured imposed by MGS.
 It is not the full \textit{process tree}  
\cite{leemans2013discovering}, but it only captures the concurrency 
conflicts that are of our interest. 
The C-tree captures the nested-concurrent 
structuring in the net. Fig. \ref{fig:ctree} shows an example net and its 
corresponding C-tree.

In a C-tree, places are denoted by their labels, and 
transitions in the net
are omitted. A C-tree \textit{node} is composed of all sequential members at a 
given level of branching in the 
concurrency hierarchy of the net. A tuple in MGS is represented in the C-tree by a box-
branching element. To elaborate,
the presence of an 
inner AND-block in a node is represented through an element denoted by a 
\textit{box} symbol $\Box$, which is  always  a member of a node. A box thus 
represents an inner AND-block, which is 
in sequence with the other places listed in that node. These inner AND-blocks 
(concurrent blocks)
are hereafter referred to as \textit{C-blocks}. Multiple  C-blocks in a single 
node indicates the presence of  multiple AND-blocks in sequence (Fig. 
\ref{fig:homomorph}(b)). The fan-out of a C-block represents the number of 
parallel branches in the corresponding AND-block. A place resides in only one 
node since there are no duplicates. A concurrent place in a net belongs to a 
non-root node in the C-tree.  
Depth in C-tree captures the concurrency hierarchy level. In the figure, $p_6$ 
is in the second inner concurrent block in the net, whereas $p_{12}$ is not in 
any concurrent block. Next, we describe how concurrent places w.r.t. a given 
place are identified from a C-tree.

\subsection{\textbf{Generator of Concurrent Submarking (GCS)}}
GCS of a given place in a net is  a C-tree structure identifying only the 
branches concurrent to that place. 

Concurrent place is defined as follows:

{\begin{defi}\label{def:concurrentPlace}
{\bf Concurrent place:}
If place p is part of a marking M, then every other place q in the same marking M is said to be concurrent to p.
\end{defi}

In the C-tree structure, if places $p$ and $q$ do not reside in the same node, and if from place $p$ to place $q$ cannot be reached by traversing direct parent-child links, they are concurrent to each other. 

The GCS is used to find potential migration 
conflicts w.r.t. a given place $p$.
\textit{GCS(p, t)} for place $p$ in C-tree $t$ 
is obtained by removing from $t$ all parts not concurrent with $p$. Links 
leading to concurrent C-tree branches are preserved.
As an example, Fig. \ref{fig:ctree} shows a net and the GCS of place $p_{6}$ in 
the net.

\subsection{\textbf{C-tree Embeds all Markings }}
A marking $M$ involving a given place $p$ can be generated using GCS($p,C$) by  
Algorithm \ref{algo:marking}, 
where $C$ is the whole C-tree to which $p$ belongs. 
The algorithm starts with the C-tree of the entire net, and first computes the 
GCS of $p$, which does not contain $p$. Then it picks up a place $q$ from it, 
and further recursively picks up the next concurrent place by computing the GCS 
of $q$. The algorithm terminates when no more concurrent place can be found. 
Though in our structural approach to compute change regions we do not generate 
individual markings, the algorithm is given to show the completeness of  the 
C-tree structure w.r.t. markings, by generating all possible markings from it. 
Function \texttt{empty(G)} checks if the C-tree $G$ does not have any place, 
i.e. it 
returns {\tt true} when
\texttt{places(G)}=$\emptyset$.

\setlength{\intextsep}{5pt}
\begin{algorithm}
 \KwIn{place $p$, C-tree $C$ of net $N$}
 \KwOut{A reachable marking $M$ in $N$ that involves $p$}

 $M\leftarrow \{p\}$;
 $G\leftarrow GCS(p,C)$\;
 \While{ {\tt not empty($G$)}    } {
    pick a place $q$ from $G$; $M\leftarrow$ $M\cup\{q\}$;
    $G\leftarrow GCS(q,G)$\;
 }
 \Return{$M$}
 \caption{Generating a Marking from a C-tree}
 \label{algo:marking}
\end{algorithm}

\subsection{\textbf{Dysfunctional C-Tree and Break-off Set}}\label{dystree}
A dysfunctional C-tree is a C-tree that cannot generate any valid reachable marking.  The purpose defining dysfunctional tree is to define break-off set, which contributes in the change region.
If after removal of some places from a C-tree, the 
 mutated C-tree cannot generate  a marking that can be generated from the 
original C-tree, the  mutated C-tree
is called {\em dysfunctional}, and the set of 
places removal of which renders the tree dysfunctional is called as a {\em 
break-off set} for that C-tree. It may be noted that the removal of places is only for demonstrative purpose, to explain what a dysfunctional tree is conceptually (to construct one such). Places are not removed in any of the algorithms. Many break-off sets may exist for a given 
C-tree. 
A dysfunctional tree
can be identified easily by presence of an empty path in the tree covering 
nodes 
without places starting from the root and reaching till a leaf.  
Such an empty path is connected  through nodes having only C-block ($\Box$) 
elements and ends at an empty leaf node. Fig. \ref{fig:ctree} shows an example of ECWS net and  its C-tree, of which a break-off set
  along with the corresponding dysfunctional C-tree is shown in Fig. \ref{fig:dys}.

\begin{figure}[t]
 \centering
 \includegraphics[scale=0.4]{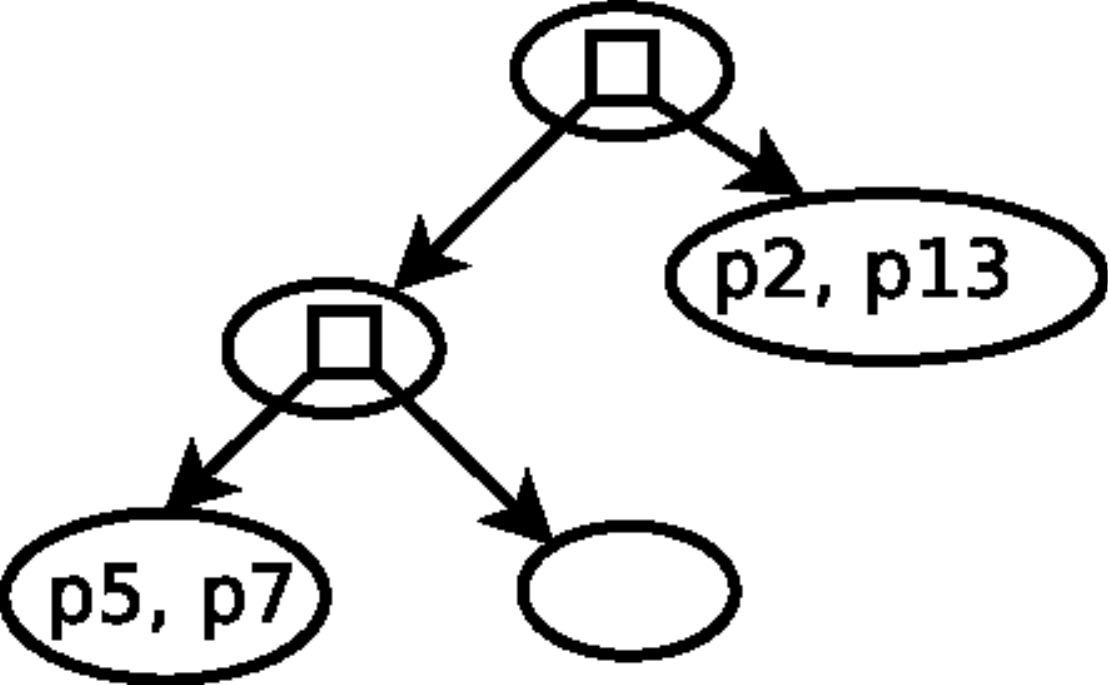}
 \caption{Dysfunctional C-tree and Break-off set 
\{$p_1,p_2$,$p_3$,$p_4$,$p_6$,$p_8$,$p_9$,$p_{12}$\} w.r.t. 
Fig. \ref{fig:ctree}}
 \label{fig:dys}
\end{figure}

\begin{figure*}[h!bt]
 \centering
 \includegraphics[scale=0.42]{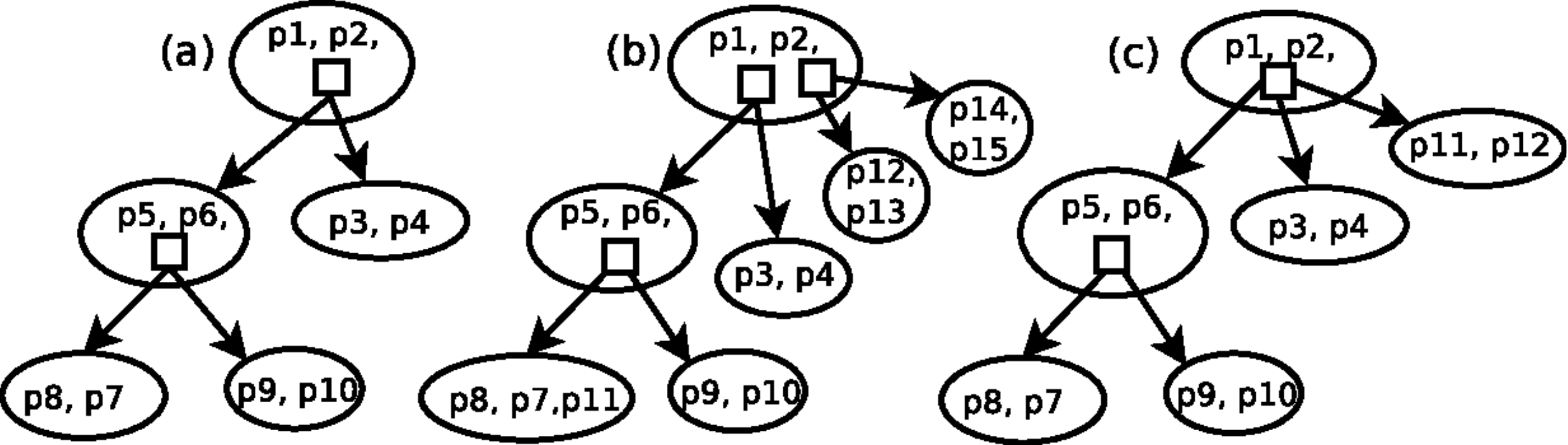}
 \caption{C-tree Marking Preserving Embedding Example}
 \label{fig:homomorph}
\end{figure*}

\subsection{\textbf{Marking Preserving Embedding (MPE) of a 
C-tree}}\label{mpe}

Let the old and the new C-trees be $C$ and $C'$ respectively. An MPE of C-tree $C$ in C-tree $C'$ is a mapping from $C$ to $C'$ 
defined recursively:
(i)  \texttt{places(root($C$))} $\subseteq$ \texttt{places(root($C'$))}, and 
(ii) C-blocks in \texttt{root($C$)} are injectively mapped to C-blocks in 
\texttt{root($C'$)} 
such 
that  within each  C-block to C-block mapping pair $(b,b')$, there is a 
\textit{bijective} MPE of the children C-trees of $b$ to those of $b'$.

For example, the C-tree in  Fig. \ref{fig:homomorph}(a) has its MPE in the 
C-tree in Fig. \ref{fig:homomorph}(b). The root nodes satisfy condition (i), 
and 
condition (ii) is satisfied as follows: the only C-block  of the root in figure 
(a) can be mapped to the left-most C-block of the root in figure (b). 
Recursively the embeddings can be observed.
On the other hand, such a mapping cannot be created between C-trees in Figs. 
\ref{fig:homomorph} (a) and (c) due to the change in fan-out of the C-block in 
the root node itself.

\begin{theo}\label{markembed}
\textbf{(Non-migratability Lemma)} {\em Given two C-trees $C$ and $C'$, if an 
MPE of $C$   
in  $C'$ does not exists, then at least one marking 
constructible from $C$ cannot be   
 constructed from $C'$. The converse is also true}.
 \end{theo}
{\em Proof:} 
If such an embedding doesn't exist, then at least one of the two constructor 
conditions is violated. If condition (i) is violated, a non-concurrent marking 
of $C$ is not available in $C'$. Then this marking is one such non-migratable 
marking.  On the other hand, if condition (ii) is violated, at least for one of 
the concurrent branches in $C$, there is no mapping to $C'$  through the nested 
structure without violating the subset condition (i).  We prove the converse by 
proving that if MPE exists, then all markings constructible by $C$ are 
constructible by $C'$. Given the existence of MPE, a non-migratable marking is 
available only if some place gets removed or some places get added from/into 
one 
of the markings. The former is not possible due to the subset condition (i), 
and 
the latter is not possible due to bijection condition (ii) in the above 
definition of MPE. Hence the proof. 

\begin{table*}[h!bt]
 \centering
 \caption{Effect of Concurrency of a place on Migratability}
 \begin{tabular}{|c ||c|c |c|}
 \hline
  Case id& 
  \textit{Concurrent in 
$N$}&\textit{Concurrent in $N'$}& Migratability\\\hline
    \hline
 (C1)&$\checkmark$ & $\checkmark$ & Conditionally migratable \\
 
 (C2) &$\checkmark$ & $\times$ & Non-migratable\\
 (C3) &$\times$&$\checkmark$  & Non-migratable\\
 (C4)& $\times$& $\times$ & Migratable\\
  (C5)&don't care & absent in $N'$& Non-migratable\\
\hline
 \end{tabular}
\label{tab:conc}
\end{table*}

\section{Change Properties of 
Places}\label{funcsigs}\label{propsection}

Let $p \in P$ be a place in the old net $N = (P, T, F)$. If $p$ is a member of 
a concurrent block (AND), all its peer concurrent places $p', p'',$ $ \dots \in 
P$ are  involved along with $p$ in some or the other markings (by Definition \ref{def:concurrentPlace}). 
If $p$ is not a part of any concurrent block, it creates only a  standalone 
marking $\{p\}$.  This case covers for $p$ in an XOR, a sequence or a loop 
branch, which is not nested inside an concurrent block.
Speaking of the new net $N'$, 
 $p$ may or may not be dropped from it. If $p$ is dropped, then marking $\{p\}$ 
is not migratable. Otherwise, $p$  may change its concurrency in $N'$. All 
markings involving a concurrent place that becomes  non-concurrent or 
vice-versa 
are non-migratable. If $p$ is non-concurrent in both the nets, then the only 
marking $\{p\}$ involving place $p$ is migratable. If $p$ is concurrent in both 
the nets, the migratability of markings involving $p$ depends on the changes to 
concurrency.
  These possibilities are summarized in
Table \ref{tab:conc}.

The non-migratability arising in cases (C1), (C2), (C3) and (C5) is attributed 
to one or more reasons (change properties) described below. The change 
properties are defined w.r.t.  place $p$ in terms of markings, and also in 
terms 
of their corresponding structural characterizations on C-trees.
Justification for the equivalence between the marking based and structure based 
formulations are also mentioned wherever it is not obvious.

For the following definitions, let $N$ and $N'$ be old and new nets and $C$ and 
$C'$ be their corresponding C-trees. 
Numbering of the definitions correspond to the cases in Table \ref{tab:conc}.
A concurrent place in a net belongs to a 
non-root node in the C-tree. Depth in C-tree represents hierarchy of 
concurrency 
in the net.

\begin{itemize}
\item 
(C5) {\bf Removal} No marking involving $p$ is reachable in $N'$. Structurally, 
$p$ is present in $C$, $p$ is absent in $C'$.  
 \item 
(C2) {\bf Lost  Concurrency}  Markings involving $p$ are concurrent in $N$ but 
not in $N'$. So, structurally,   $p$ in a non-root node in 
$C$ but in the root node in $C'$.
 \item 
(C3) {\bf  Acquired  Concurrency}  Only one standalone marking involving $p$ in 
$N$, but concurrent markings in $N'$. 
Therefore, structurally,   $p$ is in  root node of $C$ but in a non-root node 
in 
$C'$.

 \item 
(C1.1) {\bf Weak  Reformed  Concurrency} (i) $p$ is  in concurrent markings in 
both
$N$ and $N'$ and (ii) at least one concurrent marking involving $p$ in $N$ is 
not 
reachable in $N'$ due to addition or reduction of concurrency of $p$. 
Structurally, the same can be stated as:
(i) $p$ in non-root nodes in both $C$ and $C'$, 
 (ii) $GCS(p,C)$ does not have an MPE in $GCS(p,C')$.
 
 \item 
(C1.2) {\bf  Strong  Reformed Concurrency} (i) $p$ is in concurrent markings in 
$N$ and $N'$  (ii) all concurrent markings involving $p$ in $N$ are not 
reachable in $N'$. 
In terms of the structure, the same conditions are stated as: (i)
$p$ in non-root nodes both in $C$ and $C'$, and 
(ii) either the set of places  {\tt \{places}(GCS($p, C$)) $-$ 
{\tt places}(GCS($p$, $C'$))\} is a break-off set w.r.t. C-tree $GCS(p,C)$, or 
the set of places {\tt \{places(GCS($p, C'$)) $-$ 
places(GCS($p,C$))\}} is a break-off set w.r.t. C-tree $GCS(p,C')$.
\newline 
\textit{Justification}: $p$ in non-root nodes implies involvement in 
concurrency. When condition (ii) is satisfied, it
means that the places which are concurrent to $p$ in both nets are not capable 
of generating any common valid marking involving $p$. Consequently, no marking 
involving $p$ is migratable.
\end{itemize}

\begin{figure*}[h!bt]
 \centering
 \includegraphics[scale=0.35]{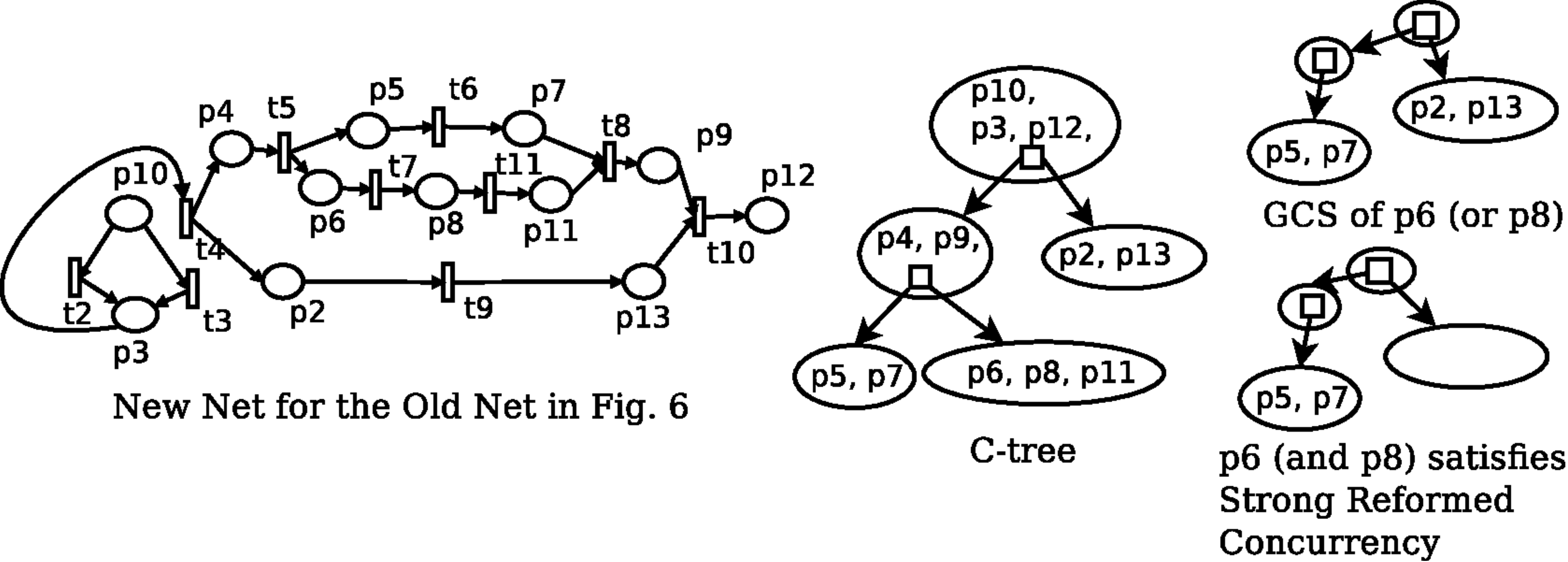}
 \caption{Verify Structural Change Properties  w.r.t. 
Fig. \ref{fig:ctree}}
\label{fig:PropertyShow}
\end{figure*}

\noindent{\bf Examples of Change Properties:} 
W.r.t. the new net in Fig. \ref{fig:PropertyShow} and an old net in 
Fig. \ref{fig:ctree}, the places in the old net are characterized as follows: 
$p_1$ satisfies removal since it is present in the old C-tree but not in the new. Place, 
$p_{10}$ satisfies lost concurrency, since in the old C-tree it belonged to a non-root node whereas in the new C-tree it resides in the root node. Similarily, since $p_2$ belonged to
the root node in the old C-tree and belongs to a non-root node in the new C-tree, it satisfies acquired concurrency. 
Further, by comparing the C-trees for both the nets it can be observed that all 
of $p_4$, $p_9$, $p_5$, $p_7$, $p_6$ and $p_8$ satisfy weak reformed 
concurrency 
since they do not have concurrent place $p_{10}$ now in the new net. $p_{11}$ 
also satisfies the same property since it does not have concurrent places $p_6$ 
and $p_8$ any more.
Lastly, Fig. \ref{fig:PropertyShow} shows the remainder of the GCS of 
$p_6$ (or $p_8$) for the old net after deleting the places in the GCS of $p_6$ 
(or $p_8$) for new net, which is a dysfunctional C-tree w.r.t. the old GCS of 
$p_6$. Thus, both
$p_6$ and $p_8$ satisfy strong reformed concurrency.

\section{Member Properties of Places}\label{memprop}
Member properties define the type of relationship of a place with the change 
region.
Every place involved in a non-migratable marking in a given old net is  
either a \textit{perfect member} or an \textit{overestimation}. The remaining 
places are \textit{safe} places in the net. These mutually exclusive types of 
memberships are now defined.

\begin{defi}\label{perfect}
 {\bf Perfect Member (of Change Region):} A place $p$ in the old net $N$ is a 
perfect member 
in $N$, w.r.t. the new net 
$N'$ iff all markings in $N$ 
involving $p$ are non-migratable.
\end{defi}

\begin{defi}\label{overestimat}
 {\bf Overestimation:} A place $p$ in the old net $N$ is an overestimation 
w.r.t. the new net $N'$, iff 
there exists a migratable marking and also a non-migratable marking involving 
$p$ in $N$.
\end{defi}

\begin{defi}\label{safe}
 {\bf Safe Member:} A place $p$ in the old net $N$ is a safe member w.r.t. the 
new net $N'$, iff 
every marking involving $p$  in $N$ is migratable.
\end{defi}

\begin{theo}\label{allmig}
\textbf{(Perfect Member Lemma)} If a place $p$ in net $N$  satisfies one 
of Removal, 
Lost Concurrency, Acquired Concurrency and Strong Reformed Concurrency w.r.t. 
net $N'$, then the place is a perfect member and vice-versa.
\end{theo}

{\em Proof:} 
We first address the sufficiency of each of the properties, which is directly 
derived from the change properties discussed in Section \ref{propsection}.  
Sufficiency of Removal follows from property (C5) that if a place $p$ in $N$, 
which 
is absent 
in $N'$ makes all markings involving $p$ in $N$ non-migratable. Sufficiency of 
Lost Concurrency follows from property (C2) that a concurrent place becoming 
non-concurrent makes all  its markings (which involve multiple places) 
non-migratable since the target marking is a standalone marking. Sufficiency of 
Acquired Concurrency  follows from property (C3)  that a non-concurrent place 
becoming concurrent makes its one and only standalone marking non-migratable. 
Sufficiency of Strong Reformed Concurrency follows from property (C1.2) that a 
place satisfying Strong Reformed Concurrency contributes only to concurrent 
markings, and all of them are non-migratable.

Now we are left 
to prove that, there is no other way for a place to 
be a perfect member. Assuming the contrary, let $p$ be a place in $N$, 
which violates all these 
properties and yet is a perfect member. Now, since 
$p$ violates the Removal property, $p$ is present in both $N$ and $N'$. When 
$p$ 
is present in the new net, and as $p$ does not satisfy  
 Lost Concurrency, we have  either $p$  not inside any concurrent block in both 
nets, or it is a part of concurrent blocks in both nets, or that it is not part 
of a concurrent block in the old net but it is so in the new net. In the first 
case, $p$ is migratable being standalone and reachable, therefore, it can 
not be a perfect member. The third case is a contradiction since we assumed 
that 
$p$ violates Acquired Concurrency. As per the assumption, since $p$ violates 
Strong Reformed Concurrency as well, in the second case, $p$ is involved in 
at least one migratable marking, implying that $p$ is not a perfect member. 
Hence the proof.

\begin{theo}\label{ultaweak}
\textbf{(Overestimation Lemma)} If a place $p$ in net $N$ satisfies Weak 
Reformed 
Concurrency but not Strong Reformed Concurrency w.r.t. net $N'$, it is an 
overestimation w.r.t. $N'$ and the vice-versa.
\end{theo} 

{\em Proof:}  
If a place satisfies Weak Reformed Concurrency but not Strong Reformed 
Concurrency, it has at least one migratable and at least one non-migratable 
marking, which is Def. \ref{overestimat} of overestimation, which proves the if 
part.

Now, for the converse argument, as per Def. \ref{overestimat}, an 
overestimation 
contributes to at least one non-migratable and at least one migratable marking. 
This implies that the overestimated place is concurrent in the old net since it 
involves in at least these two markings. Also, 
since one of the markings is migratable, the place is concurrent in the new 
net. 
This satisfies property (C1.1-i). Also, since the place is involved in one 
non-migratable marking, it satisfies property (C1.1-ii). Properties (C1.1-i) 
and 
(C1.1-ii) together ensure Weak Reformed Concurrency. Also, since the place is 
involved in at least one migratable marking, it violates Strong Reformed 
Concurrency. Hence the proof.

\section{Region Properties}\label{minimality}

A \textit{Perfect Structural Change Region} (PSCR) 
is a set of places which together cover all of the non-migratable 
markings in a net by ensuring that every non-migratable marking involves at 
least one token inside PSCR.  In addition, PSCR does not include a place 
involved in a migratable marking. If PSCR exists in a net, there is no 
overestimation of non-migratability. However, they may not exist for certain 
nets, in which case, an overestimation is unavoidable in 
the structural approach. 
All of the earlier versions of change region definitions in the literature 
known 
to us
consider a change region as a subnet.
On the contrary, the following definition characterizes a 
change region as only a set of places. 
First, a weaker change region called {\em Structural Change Region (SCR)} is 
defined, following which the notion of {\em Perfect Structural Change Region} 
(PSCR)
is introduced. In the definitions,  $M_0$ and $M'_0$ are  respectively the 
initial markings of the old net $N = (P, T, F)$ and the new net $N'=(P', T', 
F')$. Set $\mathbb{R}(m)$ represents the set of reachable markings from a given 
marking $m$ w.r.t. a given net.

\begin{defi}\label{scrdef}
 {\bf Structural Change Region (SCR)}
  Given a migration net pair consisting the old net $N$, and the new 
net $N'$, the Structural Change Region  $SCR(N,N') 
\subseteq P$ is a subset of places in $N$, such that
 $ p \in SCR(N,N')$ $\Leftrightarrow$ 
$\exists M \in \mathbb{R}(M_0)$ in $N$,
$p\in M$, $M \not\in \mathbb{R}(M'_0)$ in $N'$.
\end{defi}

\begin{defi}\label{pscrdef}
 {\bf Perfect Structural Change Region (PSCR)}
  Given a migration net pair of old net $N$, and new 
net $N'$, the Perfect Structural Change Region  $PSCR(N,N') 
\subseteq P$ is a subset of places in $N$, such that
 (i)  $ M \in \mathbb{R}(M_0)$ in $N$,  $M\not\in \mathbb{R}(M'_0)$ in $N'$ 
$\Rightarrow$ $\exists p\in M, p\in PSCR(N,N')$, and  (ii) $p \in PSCR(N,N')$ 
$\Rightarrow$ 
$\forall M\in \mathbb{R}(M_0)$ in $N$ s.t. $p\in M$, 
$M\not\in \mathbb{R}(M'_0)$ in $N'$.
\end{defi}

It can be observed that Definition \ref{scrdef} uses operator $\Leftrightarrow$ to establish relation between a place that is a member of the change region and its contribution to non-migratability. If we break down the definition using $\Rightarrow$ operator (as used in Definition \ref{pscrdef}), the following two conditions arise: 
(i) $M \in \mathbb{R}(M_0)$ in $N$, $M\not\in \mathbb{R}(M'_0)$ in $N'$ $\Rightarrow$ $\exists p\in M$, $p \in SCR(N,N')$.
(ii) $p \in SCR(N,N')$ $\Rightarrow$ $\exists M \in \mathbb{R}(M_0)$ in $N$, $p\in M$, $M\not\in \mathbb{R}(M'_0)$ in $N'$. Condition (ii) in the above is different from the condition (ii) of Definition \ref{pscrdef} due to use of the existential quantifier used in the RHS of the implication, which makes
SCR is a weaker version of PSCR. SCR holds all those places for each of which 
there is a non-migratable marking, but it permits overestimation. On the other 
hand, if an SCR does not have 
any migratable marking it becomes PSCR. $PSCR(N,N')$ is a partial function on 
$N$ and $N'$. It may not exist on certain cases. Non-existence of PSCR implies there is non-migratability between the given net pair but there is no minimal change region covering it. In other words, no set can be found which satisfy Definition \ref{pscrdef}. The example in Figure \ref{fig:overunder2}(a)-(b) elaborates this case. On the other hand, PSCR may exist as empty 
set 
for certain cases indicating full migratability of the markings of the old net. In this case, set $\phi$ satisfies Definition \ref{pscrdef}.
If it has member places, they together cover all non-migratable markings.
Condition (i) of PSCR  ensures that if a marking in $N$ is non-migratable, it 
always includes a 
member place from PSCR.
 Condition (ii) of PSCR ensures that if a place $p$ is inside PSCR, then no 
reachable 
marking in the old net involving $p$  is reachable in the new net. 
  PSCR eliminates the overestimation (false-negatives) occurring inside change 
regions as computed by the previous approaches.  Also,  false-positives outside 
the PSCR are eliminated.
Condition (i) is the baseline property of structural reachability based change 
regions, which is 
satisfied by previous versions of change regions in the literature discussed 
above. 
SESE-based approaches cause overestimations for nets shown in Figs. 
\ref{fig:andvdamore} and \ref{fig:loopxorcr}, though PSCRs 
eliminating overestimations exist for them.  In these cases of full 
migratability the PSCRs happen to be empty.

We have noted earlier that not all nets have PSCRs. If the PSCR exists for a 
net, the members of PSCR gives the set of all {\em perfect members} in the net. 
On the other hand, in absence of PSCR in a given net, non-migratability is 
captured by places in the union set of all {\em overestimations} and  
all perfect members, which represents the SCR, as stated through the following 
Lemma. 
\begin{theo}\label{scrLemma}
 \textbf{(SCR Lemma)} The union of all overestimations and all perfect 
members in net $N$ 
w.r.t. net $N'$ is the SCR in $N$ w.r.t. $N'$.
\end{theo}

{\em Proof}: Every non-migratable marking includes at least a place from the 
SCR (Def. 
\ref{scrdef}). Also,  overestimation and perfect members (Defs. 
\ref{overestimat},\ref{perfect}) together cover all non-migratable markings. 
Hence the proof.\\

In absence of 
PSCR, the SCR can be used  change region. 
However, it may be further optimized by removing those overestimated places 
which are fully covered by at least one other place. 
Now, we formulate the condition for existence of PSCR for a given net 
pair. 

\begin{theo}\label{exist} 
\textbf{(PSCR Lemma)} PSCR exists in a given net $N$ 
w.r.t. some net 
$N'$ iff every non-migratable marking in $N$ includes at least one perfect 
member.
\end{theo}

\textit{Proof}: 
Given that every non-migratable marking includes at least one perfect member, 
the set of all perfect members constructs PSCR by satisfying Def. \ref{perfect}. 
 Hence, by definition, if every non-migratable marking includes at least one 
perfect 
member, PSCR exists.
On the other hand, if PSCR exists, every non-migratable marking includes some 
place from 
PSCR (Def. \ref{perfect}(i)). Being a member of PSCR, that place is involved in 
no migratable marking (Def. \ref{perfect}(ii)), which makes that place a 
perfect 
member. Thus, if PSCR exists, every non-migratable marking involves at least one 
perfect member. Hence the proof.\\

\begin{table*}[h!bt]
\caption{Computation-friendly  Cases for Existence of PSCR}
\begin{tabular}{|p{2cm}|p{3.5cm} |p{3.5cm}| p{4cm}|p{1cm}|}
\hline
Are there overestimations? ($Over \neq \emptyset$) &There are markings in $C$ 
without perfect members (i.e. $Perf$ is not break-off set for $C$)&There are 
markings in $C'$  without perfect 
members (i.e. $Perf$ is not
break-off set for $C'$)&All markings in $C$  without a perfect member are valid 
in $C'$(i.e. {\tt delete}($C$,$Perf$) has MPE in 
{\tt delete}($C'$,$Perf$)) & 
PSCR exists\\
\hline\hline

$\times$&don't care& don't care& don't care& $\checkmark$\\
\hline
$\checkmark$&$\times$ &don't care & don't care& 
$\checkmark$\\\hline
$\checkmark$& $\checkmark$  & $\times$  & don't care & $\times$\\\hline

$\checkmark$& $\checkmark$  & $\checkmark$ & $\times$  & $\times$\\\hline

$\checkmark$& $\checkmark$  &$\checkmark$ & $\checkmark$ & $\checkmark$\\\hline
\end{tabular}
\label{tab:pscrcases}
\end{table*}

\begin{figure*}[h!bt]
 \centering
 \includegraphics[scale=0.32]{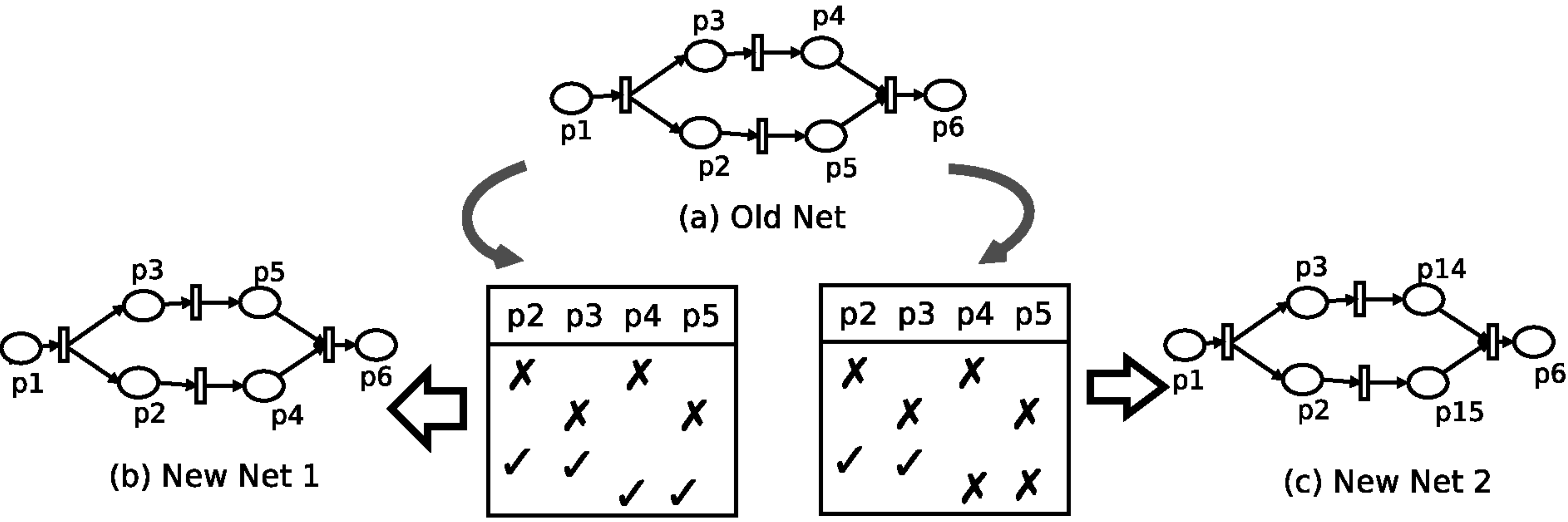}
 \caption{Migratable and Non-migratable Markings in two Migration Pairs}
\label{fig:overunder2}
\end{figure*}

\noindent{\bf Examples of Existence and Non-existence of PSCR:}  Fig. 
\ref{fig:overunder2}(a)-(b) illustrates a case 
where 
the PSCR does not 
exists, whereas, for Fig. 
\ref{fig:overunder2}(a)-(c), PSCR exists which is set $\{p_4,p_5\}$. The SCRs 
of the old net in Fig. \ref{fig:overunder2}(a) w.r.t. the new nets in Figs. 
\ref{fig:overunder2}(b) and (c)  are $\{p_2,p_3,p_4,p_5\}$ and 
$\{p_2,p_3,p_4,p_5\}$ respectively.
 Migratability of the markings inside the AND block of the old net for these 
two 
cases is
captured through two tables in the figure. A row in a table corresponds to a marking in the net (a). Symbols $\times$  and $\checkmark$  
respectively denote a token in a non-migratable marking, and a token in a migratable marking. For example, for marking \{$p_2$,$p_4$\} that is migratable from net (a) to (b), $\checkmark$ appears in the corresponding columns in the table for case (a)-(b). In case (a)-(b), PSCR 
does not exist since all places inside the concurrent block are involved in one 
migratable and one 
non-migratable marking.  In case (a)-(c), PSCR exists since perfect members 
$p_4$ and $p_5$ cover all non-migratable markings. \\

\subsection
{Computation for Existence of PSCR}\label{computeExist}
From Lemma \ref{exist}, we can see that PSCR does not exist if and only if 
there 
is at least one non-migratable marking in old net that does not include any 
perfect member.
Table \ref{tab:pscrcases} identifies structural conditions for the existence of 
PSCR. 	If there are no overestimations, since there cannot be any 
non-migratable marking not including a perfect member, PSCR exists. This case is 
covered in row 1. The remaining rows in the table cover the cases of presence of 
overestimation.
If there are no markings in $C$ without involving perfect members, then 
PSCR exists (row 2). Structurally, the perfect member set is a break-off set in 
$C$. In the remaining three cases, $C$ contains some markings without involving 
perfect members. In addition, if all markings in $C'$ have perfect members (row 
3), the markings without perfect members in $C$ are not migratable, and 
therefore, PSCR does not exist. Structurally, perfect members form a break-off 
set for $C'$ but not for $C$. Instead if $C'$ also has some markings without 
perfect members (row 4, 5), then we want to check whether all markings of $C$ 
without perfect members can be generated from $C'$. Structurally, this can be 
done by first removing the set of perfect members from both the C-trees, and 
then checking if the mutated $C$ is embedded in the mutated $C'$. \\

\subsection{Computing SCR and PSCR}\label{computeCRs}

After the discussion on  SCR and PSCR in terms of perfect members and 
overestimations, we now connect with them the five structural (C-tree based) 
change properties discussed in Section \ref{propsection} through the following 
{\em Perfect Member} Lemma, to summarize 
computation of SCR and PSCR as shown in Table \ref{algo:algo}.
The corresponding algorithm to compute SCR and PSCR is listed in Algorithm \ref{algo:scrpscr}. 

{\color{red}
\setlength{\intextsep}{5pt}
\begin{algorithm}
 \KwIn{place $p$, C-tree $C$ of net $N$}
 \KwOut{A reachable marking $M$ in $N$ that involves $p$}

 $M\leftarrow \{p\}$;
 $G\leftarrow GCS(p,C)$\;
 \While{ {\tt not empty($G$)}    } {
    pick a place $q$ from $G$; $M\leftarrow$ $M\cup\{q\}$;
    $G\leftarrow GCS(q,G)$\;
 }
 \Return{$M$}
 \caption{Generating a Marking from a C-tree}
 \label{algo:scrpscr}
\end{algorithm}
}

\setlength{\intextsep}{5pt}
\begin{table*}
\centering
  \caption{Computation of SCR, PSCR and other sets}

\begin{tabular}{|l|p{7cm}|p{2.5cm}|}
\hline
{\em Sets} & {\em Set Constructors \newline ($C$ and $C'$ are the C-trees of the 
old and 
the new net)}& {\em Reference Definitions, Lemmas} \\\hline\hline
 $CR_r$ &{\tt removal($C$, $C'$)}&Section \ref{funcsigs}(C5)\\
 $CR_{lc}$ & {\tt lostConcurrency($C$, $C'$)}&Section \ref{funcsigs}(C2)\\
  $CR_{ac}$ & {\tt acquiredConcurrency($C$, $C'$)}&Section \ref{funcsigs}(C3)\\
   $CR_{wrc}$ & {\tt weakReformedConcuurency($C$, $C'$)}& Section 
\ref{funcsigs}(C1.1)\\
   $CR_{src}$ & {\tt strongReformedConcurency($CR_{wrc}$, $C$, $C'$)}&
Section \ref{funcsigs}(C1.2)\\\hline
    $Over$& $CR_{wrc} - CR_{src}$& Lemma \ref{ultaweak}\\\hline
    $Perf$& $CR_r \cup CR_{lc} \cup CR_{ac} \cup CR_{src}$ & Lemma 
\ref{allmig}\\\hline
    $SCR$ & $Over \cup Perf$& Lemma \ref{scrLemma}\\\hline
  $PSCR$ & $Perf$ iff {\tt PSCRexists($C$, $C'$, $Over$, $Perf$)}& Lemma 
\ref{exist}\\\hline 
    
\end{tabular}
 \label{algo:algo}
\end{table*}

\begin{figure*}[h!bt]
\centering
\includegraphics[scale=0.35]{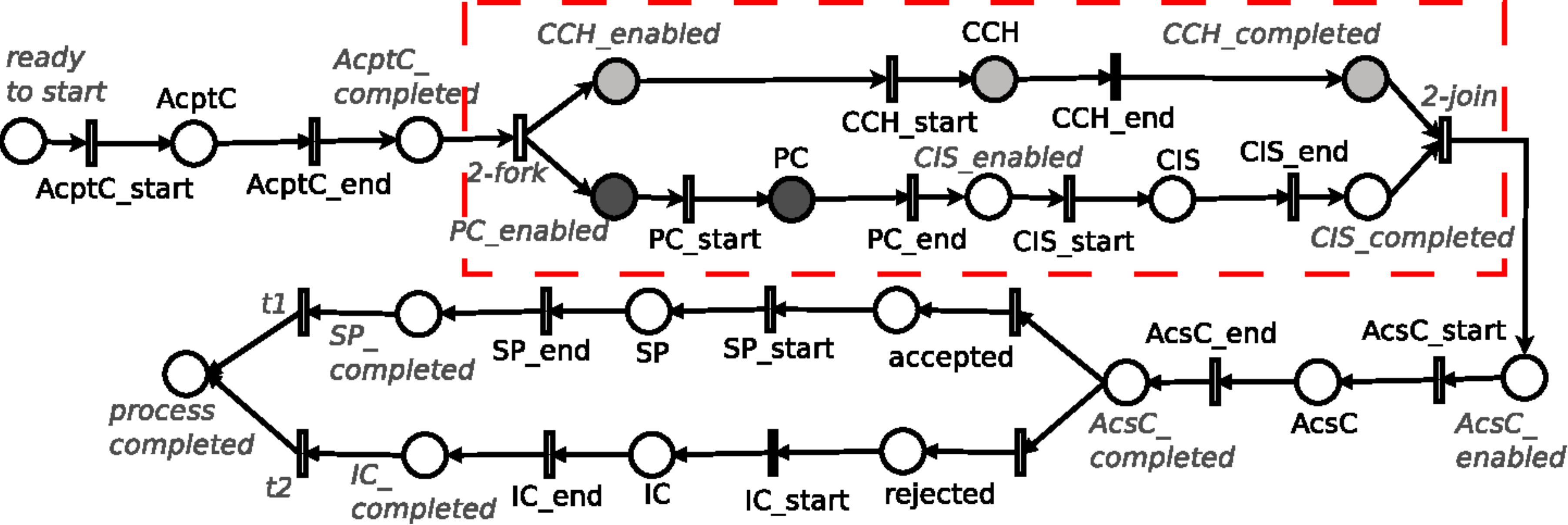}\\
(a) Old Process\\
\vspace{2mm}
\includegraphics[scale=0.35]{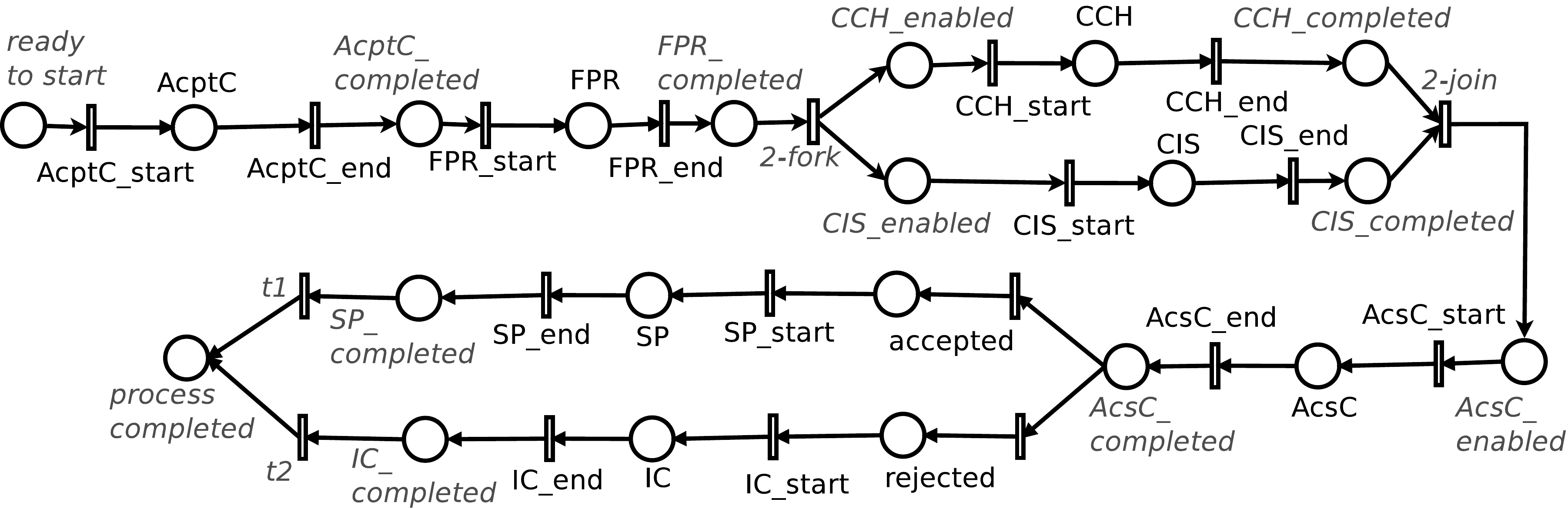}\\
(b) New Process\\
\caption{WF-net models of some BPMN models given in Hens et al. 
\cite{hens2013process} (Claims process)}
\label{fig:hens}
\end{figure*}
 
SCR,  PSCR and other sets listed in the table are computed from the two C-tree 
structures corresponding to the two nets. 
The first row in the table provides constructors for sets of places satisfying 
the five change properties.
These are (i) places which are removed ($CR_r$), (ii) places which loose their 
concurrency ($CR_{lc}$), (iii) sequential places which acquire concurrency 
($CR_{ac}$), (iv) concurrent places for which concurrency has  changed partially 
($CR_{wrc}$) (iv)concurrent places for which concurrency has changed fully 
($CR_{src}$).
Using these sets,  set $Over$ consisting of places that are overestimations, and 
set $Perf$ consisting of places that are
 perfect members can be constructed.  
 Constructors for sets
 SCR and PSCR  are shown  in the last two rows.
  Set constructors in the table are either shown as functions on the two 
C-trees, or as set operations. The table also lists cross references to the 
properties discussed previously 
that  build the sets. For example,  function {\tt lostConcurrency} is 
implemented in terms 
of structural condition mentioned in Section 
 5(C2), which is, {\em place $p$ in a non-root node in $C$, but it is in root 
node
in $C'$}.  This condition is checked for every place $p$ in $C$. The same is 
applicable for all functions used in the table referring to the change 
properties. Function {\tt PSCRexists} implements the logic given in Section 
\ref{computeExist}.

\subsection{Utility of the Approach}\label{result}

In this section, we compare the approach of the \textit{PSCR} with  
\textit{minimal 
SESE-based} change regions highlighting the improvement offered by our approach.
These 
approaches are evaluated using the following three migration net pairs:  (i) 
The 
training process discussed in Section \ref{overestimation} and shown in Fig. 
\ref{fig:cr}, (ii) The  hypothetical processes used earlier in Figs. 
\ref{fig:ctree} and \ref{fig:PropertyShow}, and (iii) A claims process given in 
the work of Hens et al. \cite{hens2013process}, for which, Fig. \ref{fig:hens} 
shows the WF-net models translated from their original BPMN 
models.  
The dashed region in 
Fig.\ref{fig:hens}(a) depicts the minimal SESE change region, whereas, the 
 places which are shaded cover the SCR.  The PSCR is marked by places shaded 
dark.

Let us discuss the advantage of using PSCR over the minimal SESE approach 
through the example of training process in Fig. \ref{fig:cr}. Since the minimal 
SESE region is much bigger than the PSCR, it includes many false-negatives. For 
instance, the upper XOR-branch (staring with transition $s_{t_2}$, ending with 
transition $e_{t_6}$) is in the fold of such false-negatives. As a result, in 
this particular process, the 
old marketing trainees are not allowed to migrate into the new process, whereas, 
we can observe that it is safe to let them migrate as per the business goal. 
 The corresponding markings are identified as non-migratable in 
the minimal SESE approach contributing to the false-negatives. 
The PSCR approach excludes such false-negatives, confines its change region to 
the minimal set of places thereby eliminating unnecessary delay 
in  immediate dynamic migration.

\begin{figure}[h!bt]
\centering

\includegraphics[scale=0.6]{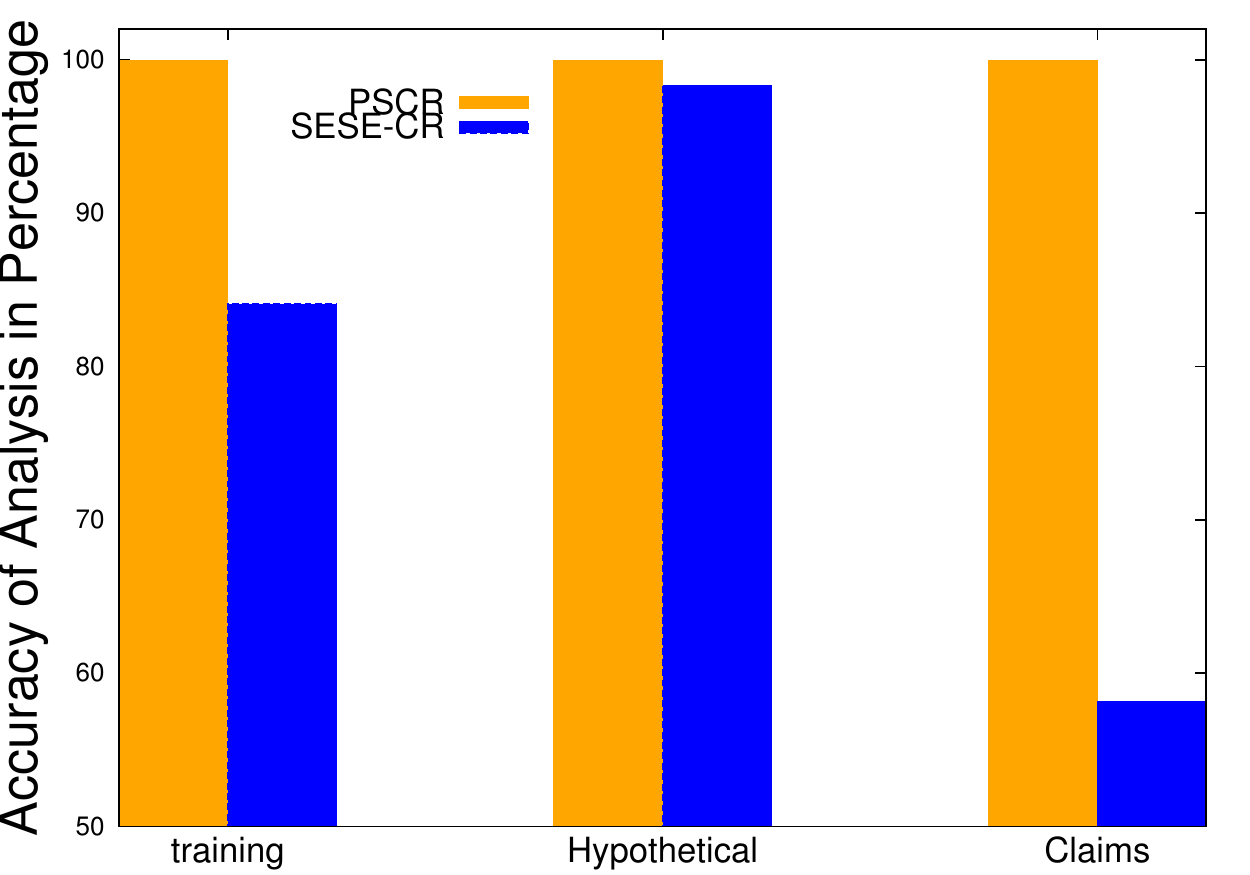}

\caption{Comparing Approaches to Deciphering Migratability}
\label{fig:plots}
\end{figure}

The comparison results from the three example processes are given in Fig. 
\ref{fig:plots}. Given the set of all possible markings in the old net, we 
compare the results of their migratability. The set of old 
markings is assumed to be available. Change regions are pre-computed separately 
by applying PSCR and minimal SESE approaches.  Then for the 
accuracy test,  each of 
the markings are verified through the change regions. 
However, since 
the minimal SESE-based approach suffers from overestimation, it has lower 
accuracy, whereas the accuracy of PSCR is 100\%. 

When PSCR does not exist, SCR 
may be used. Our recent work on distributed computation of change regions 
\cite{DBLP:conf/apn/PradhanJ16} use an 
optimized version of SCR. However, the result of using SCR may cause as many 
false-negatives 
as the minimal SESE approach in the worst case. Fig. \ref{fig:hens} represents 
such a 
case. Here, the PSCR consists of only two places, $PC_{enabled}$ and 
$PC$. However,  in 
addition, the SCR contains all the places of the upper AND-branch(lightly 
shaded). As a result, every possible marking in the AND-block 
is non-migratable as per the SCR, and similar result is reached by the minimal 
SESE approach. In this case, the PSCR is to be used since it is devoid of 
false-negatives.

\section{Conclusions}
The paper addressed the problem sufficiency of structural change regions 
in dynamic process migration. The 
issue 
of inclusion of false-negatives in 
the existing computation methods of structural change regions  was brought up. 
The reason for false-negatives in the structural approaches was localized to 
overestimation of non-migratability. 
We 
formulated the notions of SCR and PSCR with the help of a few Lemmas.
It is proved that the PSCR, if it exists, gives the accurate account of 
non-migratability. The 
necessary and sufficient conditions for PSCR to exist were provided. 
In a nutshell, this work touches upon the best possible harnessing of the 
structural change region based on transfer validity, and identifies various 
structural properties about its existence and computation. The work also 
establishes that the structural approach becomes 
inaccurate in computation of non-migratability  resulting in 
false-negatives due to the non-existence of PSCR. If the PSCR exists, 
it obtains an accurate result of non-migratability by a mere structural approach.  In 
absence of PSCR, SCR or its optimized form may be used. Another feature of this approach is that, it is not designed from change operations, making it orthogonal to the set of changes due to which a migration scenario arises.
Further work in this area consists of 
optimization of SCRs in absence of PSCR, algorithmic variations for structural computation of SCR 
and PSCR, and further generalization of this \textit{theory of change regions} 
to cover other consistency criteria. 

\bibliographystyle{IEEEtran}
\bibliography{ase}

\end{document}